\documentclass[twoside,11pt]{article}

\usepackage{jmlr2e}
\usepackage{amsmath}
\usepackage{upgreek}
\usepackage{algorithm}
\usepackage{algpseudocode}

\usepackage{multicol}
\usepackage{multirow}

\usepackage{subcaption}
\usepackage{cleveref}

\usepackage{enumitem}

\jmlrheading{1}{2021}{1-37}{4/21}{00/00}{meila00a}{Maysam Behmanesh, Peyman Adibi, Jocelyn Chanussot and Sayyed Mohammad Saeed Ehsani}

\ShortHeadings{Functional Mapping of Spectral Descriptors and Manifold Regularization}{Behmanesh, Adibi, Chanussot and Ehsani}
\firstpageno{1}

\begin{document}

\title{Cross-Modal and Multimodal Data Analysis Based on Functional Mapping of Spectral Descriptors and Manifold Regularization }

\author{\name Maysam Behmanesh \email mbehmanesh@eng.ui.ac.ir \\
       \addr Artificial Intelligence Department, Faculty of Computer Engineering, University of Isfahan, Iran
       \AND
       \name Peyman Adibi \email adibi@eng.ui.ac.ir \\
       \addr Artificial Intelligence Department, Faculty of Computer Engineering, University of Isfahan, Iran
       \AND
       \name Jocelyn Chanussot \email jocelyn.chanussot@gipsa-lab.grenoble-inp.fr \\
       \addr Univ. Grenoble Alpes, CNRS, Grenoble INP, GIPSA-lab, 38000 Grenoble, France
       \AND
       \name Sayyed Mohammad Saeed Ehsani \email ehsani@eng.ui.ac.ir\\
       \addr Artificial Intelligence Department, Faculty of Computer Engineering, University of Isfahan, Iran}

\editor{}

\maketitle

\begin{abstract}
Multimodal manifold modeling methods extend the spectral geometry-aware data analysis to learning from several related and complementary modalities. Most of these methods work based on two major assumptions: 1) there are the same number of homogeneous data samples in each modality, and 2) at least partial correspondences between modalities are given in advance as prior knowledge. This work proposes two new multimodal modeling methods. The first method establishes a general analyzing framework to deal with the multimodal information problem for heterogeneous data without any specific prior knowledge. For this purpose, first, we identify the localities of each manifold by extracting local descriptors via spectral graph wavelet signatures (SGWS). Then, we propose a manifold regularization framework based on the functional mapping between SGWS descriptors (FMBSD) for finding the pointwise correspondences. The second method is a manifold regularized multimodal classification based on pointwise correspondences (M$^2$CPC) used for the problem of multiclass classification of multimodal heterogeneous, which the correspondences between modalities are determined based on the FMBSD method. The experimental results of evaluating the FMBSD method on three common cross-modal retrieval datasets and evaluating the (M$^2$CPC) method on three benchmark multimodal multiclass classification datasets indicate their effectiveness and superiority over state-of-the-art methods.
\end{abstract}


\section{Introduction}
\label{1.intro}

Many real-world phenomena include several multiple heterogeneous data sources, which is referred to as multimodal data. Different realistic data modalities are characterized by various domains with probably very different statistical properties. Unlike the single-modal data that indicates the partial information of a phenomenon, multimodal data often contain complementary information from various modalities \citep{Lahat2015, Xu2015}. Although each modality has its distinct statistical properties, those are usually semantically correlated and multimodal learning methods try to discover the relationship between them.  Multimodal learning methods fuse the encoded complementary information from multiple modalities to find a latent representation of data and model all modalities simultaneously. Learning from multiple modalities enables capturing correlation between them and gives an in-depth understanding of natural phenomena \citep{Baltru2019}. \par
Multimodal manifold learning is an important category of multimodal learning methods that extends spectral or geometry-aware data analysis techniques for information fusion given multiple modalities. During last decade, datasets acquired by multimodal sensors have become increasingly available in real-world applications and various methods have been developed to analyze them considering their geometric structure in different modalities.\par
Most recent methods, as will be seen in the next section, present a framework that is limited to multimodal data with the same number of data samples in each modality. Also, in most of the recent methods, the correspondence between different modalities, such as correspondences between data samples (pointwise correspondences) or correspondences based on label information between categories of data samples (batch correspondences), is necessary for learning models, and this prior knowledge is determined in advance in these methods. Another restrictive assumption in several previous related methods, is the homogeneous assumption about data in different modalities.\par
In this paper, we emphasis on a more practical scenario in which, 1) there are heterogeneous data in each modality, 2) there may be not the same number of data samples in different modalities, and 3) pointwise and batch correspondences between modalities are unknown. It is only known that the data points of various modalities are sampled from the same phenomena.\par
In order to develop effective solutions for manifold based multimodal data processing in this scenario, here, we focus on two existing challenges in this area and try to solve them based on our new ideas, as discussed below.\par
The first challenge is to introduce a model that, while implicitly uses the semantic correspondences between data samples in various modalities, is independent of the exact correspondence information between them.\par
For intercepting this challenge, we introduce a model that first uses the spectral graph wavelet transforms \citep{HAMMOND2011} to extract the local descriptors on graph Laplacian of each modality. Then, since each wavelet transform on each modality is considered as a function on the manifold, we find a functional map \citep{Ovsjanikov2012} that tries to preserve these descriptors in mapping between various modalities. For finding these maps, we present a regularization-based framework between every two modalities that tries to find a map between local descriptors on these two modalities while satisfying two between-modality and within-modality constraints. Finally, we use this functional map for cross-modal retrieval problems to find the most similar samples in one modality to each query sample in another one.\par
The second challenge is using the above mentioned more practical scenario in multimodal multiclass semi-supervised classification. Most of the recent methods for multimodal classification supposed to have homogeneous data with the same number in each modality and are limited to a few real-world problems \citep{Minh2016}.\par
To tackle this challenge, we introduce a new multiclass multimodal method that combines the reproducing kernel Hilbert spaces (RKHS) and manifold regularization framework for the problem of classification of multimodal heterogeneous data, where the correspondences between modalities are determined based on the proposed model regarding the first challenge.\par
\begin{figure}[t!]
\centering
\includegraphics[width=\textwidth]{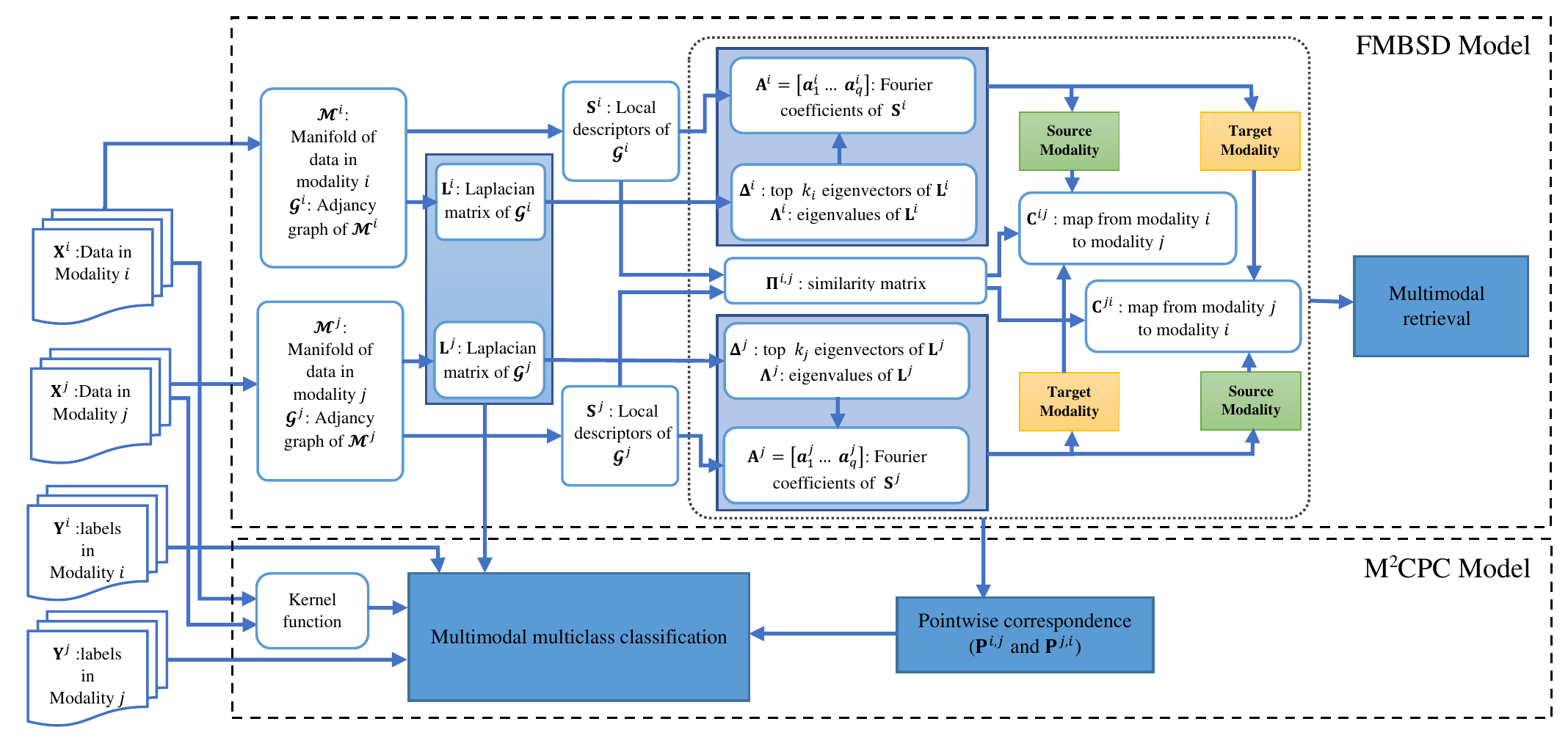}
\caption{Diagram of two proposed models and their relationships }
\label{fig1}
\end{figure}

Fig. \ref{fig1} schematically shows the diagram of two proposed models in this paper and the relationship between them. The main contributions of this paper are briefly summarized as follows:\par

\begin{enumerate}
\item We considered a general scenario for multimodal problems with unpaired data in the heterogeneous modalities, which have no prior knowledge (neither pointwise nor batch correspondences).
\item We designed a new strategy, which is not limited to homogeneous modalities with the same number of data, and is applicable for heterogeneous modalities. This strategy extracts the local descriptors of graph Laplacian of each modality using the spectral graph wavelet transform.
\item We developed a new loss function with two between-modality and within-modality regularizers that is independent of direct correspondences between samples, while implicitly uses the semantic relationship between them.
\item We introduced a new similarity measure between samples on various modalities based on local descriptors and used it for measuring between-modality similarities.
\item We developed a new manifold regularized framework for classification of multiclass multimodal data while takes advantage of the capabilities of mapping heterogeneous data in the kernel Hilbert spaces.
\end{enumerate}

The remainder of the paper is organized as follows. Related work is reviewed in section \ref{2.RelatedWork}. In section \ref{3Background} we overview the background and the basic notations. Our proposed methods are introduced in section \ref{4ProposedMethods}. Experimental results are presented and discussed in section \ref{5ExperimentalResults}. Finally, section \ref{6.Conclusions} concludes the paper.

\section{Related work}
\label{2.RelatedWork}

Some multimodal learning approaches learn a common low-dimensional latent space from different modalities. For example, variational autoencoder (VAE) neural networks are used to learn a shared latent representation of multiple modalities \citep{Yin_Huang_Gao_2020}. Several Bayesian inference methods use the ability of exploring the spatial and temporal structures of data to fuse the modalities into a joint representation \citep{Huang2013}. In another method, called MVML-LA, a common discriminative low-dimensional latent space is learned by preserving the geometric structure of the original input \citep{ZHAO2018154}. A novel model is also proposed which maps high-dimensional data to a low-dimensional space using a framework that encompass manifold learning, dimensionality reduction, and feature selection \citep{GAO2020}. Another method represents multiple modalities in a common subspace using the Gaussian process latent variable model (GPLVM) \citep{LI2021}. A learnable manifold alignment (LeMA) method is presented in \citep{HONG2019}. In this method a joint graph structure is learned from data, and data distribution is captured by graph-based label propagation. \par
Most of the above methods are limited to problems with the homogeneous data in all modalities and the same number of data samples in each of them, and are sometimes called multi-view problems. \par
Another category of multimodal learning methods tries to extend different diffusion and spectral geometry-aware methods to the multimodal setting by projecting unimodal representations together into a joint representation of multiple modalities \citep{Baltru2019}. A kernel method for manifold alignment, called KEMA, focuses on finding a mapping function to match all modalities through a joint representation and uses the traditional manifold learning techniques in the common space \citep{Tuia2016}. \par
A nonlinear multimodal data fusion method is proposed in \citep{KATZ2019}, which captures the intrinsic structure of data and relies on minimal prior model knowledge. A multimodal image registration approach is presented in \citep{ZIMMER2019} that uses the graph Laplacian to capture the intrinsic structure of data in each modality by introducing a new structure preservation criterion based on Laplacian commutativity. Another approach is presented in \citep{Eynard2015} that simultaneously diagonalizes Laplacian matrices of multiple modalities by extending spectral and diffusion geometry of them. \par
Most of these methods require to find the correspondence information between modalities because this knowledge is essential for multimodal learning from various heterogeneous feature spaces and represents the semantic relationships between them. Some of these methods, such as \citep{KATZ2019} and \cite{ZIMMER2019}, assume that the data samples in various modalities are completely paired and all modalities have the same number of data samples. Also, in some of them, such as \citep{Tuia2016} and \citep{Eynard2015}, the paired data samples in various modalities are assumed to be determined in advance using an expert manipulation process. These assumptions significantly limit the application scope of multimodal problems. \par

Recently, several methods have been introduced to reduce dependency on the expert manipulation process by extending limited amount of knowledge in the initial correspondences, using the geometry of data to find more correspondences. For example, the method presented in \citep{POURNEMAT2021} first extends the given correspondence information between modalities using functional mapping idea on the data manifolds of the respected modalities, and then uses all correspondence information to simultaneously learn the underlying low-dimensional common manifolds by aligning the manifolds of different modalities. In another recent method \citep{Behmanesh2021}, we proposed another multimodal manifold learning approach, called local signal expansion for joint diagonalization (LSEJD), which uses the intrinsic local tangent spaces for signal expansion of multimodal problems.\par

Although these two later methods greatly expand correspondence information, they are still dependent on little basic prior knowledge of correspondences.\par
The idea of finding correspondence information in multimodal problems is widely used in cross-modality information retrieval problems. These problems take query data from a modality and try to find the most relevant data from another one. Several methods have been developed to address these problems. Some of them, such as \citep{ZHANG2020}, \citep{Hong2020} and \citep{Wang2016}, try to learn a common representation of various modalities, which retrieves the relevant data using a distance measure between data in this common representation. Many cross-modal retrieval methods are based on hashing approaches, such as \citep{Tang2016}, \citep{LiuLi2017}, and \citep{WangJun2012}, which are widely used in multimedia retrieval. In these methods, heterogeneous multimodal data are embedded into a common
low-dimensional Hamming space for large-scale similarity search.\par

Similar to most of the multimodal learning methods, pointwise correspondences are vital for most of the mentioned cross-modal retrieval methods that all of the training samples of various modalities are completely paired.\par

Cross-modal retrieval problems can be divided into semi-supervised and unsupervised categories.  Unlike unsupervised approaches, such as CCA \citep{Rasiwasia2010}, PLS \citep{Sharma2011}, FSH \citep{Liu2017}, and CRE \citep{Mengqiu2019}, which take into account the geometric structure of underlying data or topological information of modalities, semi-supervised methods, such as SCM \citep{Pereira2014OnTR}, JFSSL \citep{Wang2016}, LCMFH \citep{8423193}, CSDH \citep{LiuLi2017}, SePH \citep{Zijia2015}, and DDL \citep{LIU2020199} exploit the correspondences based on label information between categories of data (batch correspondences) to improve the performance.\par

Several deep neural network-based models, such as Deep-SM \citep{Wei2017}, SCH-GAN \citep{ZhangJian2020}, CCL \citep{Peng2018}, and S$^3$CA \citep{Yang2020} have also been widely considered for cross-modal retrieval problems. Similar to most methods, they are dependent to prior knowledge, such that CCL uses pointwise correspondences information and Deep-SM, SCH-GAN, and S$^3$CA use batch correspondences information.\par

These two types of correspondences, pointwise and batch correspondences, are very strong assumptions and cannot handle in many of realistic problems. In last recent years, a few works are exploited for the practical problems, such as UCMH \citep{GAO2020178} which address the data with completely unpaired relationships by mapping data of different modalities to their respective semantic spaces, and FlexCMH \citep{9223723} which introduces a clustering-based matching strategy to find the potential correspondences between samples across modalities.

\section{Background}
\label{3Background}
The preliminary concepts needed for development of the two formulations proposed in this paper and the base models related to them, are briefly reviewed in this section. The first formulation presents a framework to find pointwise correspondences between modalities and is used for cross-modal retrieval problems. The second formulation is used for semi-supervised multiclass multimodal classification.

\subsection{Problem formulation}
\label{3.1ProblemFormulation}
We suppose that multimodal data are acquired from $m$ $(>2)$ different modalities or spaces $\left\{\mathcal{S}^i \right\}_{i=1}^m$ with different dimensions $\left\{d_i \right\}_{i=1}^m$ and data in each modality $i$ represented as a $\xi_i$-dimensional underlying manifold $\mathcal{M}^i\subset \mathbb{R}^{d_i}$ embedded into a $d_i$-dimensional Euclidean space $\mathcal{S}^i$, where $\xi_i\ll d_i$.\par

Let $\left\{\mathbf{X}^i \right\}_{i=1}^m$ are a collection of data in $m$ modalities that collected from different spaces $\left\{\mathcal{S}^i \right\}_{i=1}^m$ independently, that each $\mathbf{X}^i=[\mathbf{x}_1^i ... \mathbf{x}_{N_i}^i]^T \in \mathbb{R}^{N_i\times d_i}$ contains data in modality $i$ with dimension $d_i$, and $N_i=l_i+u_i$ is the number of samples in modality $i$, which $l_i$ and $u_i$ are the number of labeled and unlabeled samples, respectively.\par
In this paper, it is assumed that the sample correspondences information across different modalities is unknown. \par

\subsection{Data spectral geometry}
\label{3.2DataSpectralGeometry}

The weighted neighborhood graph of data for modality $i$, denoted by $\mathcal{G}^i=(\mathcal{V}^i,\mathcal{E}^i)$, is defined, where the data points are assumed to be sampled from a differentiable manifold $\mathcal{M}^i$. The graph vertices $\mathcal{V}^i=\lbrace \mathbf{x}_1^i,...,\mathbf{x}_{N_i}^i \rbrace\subset \mathcal{S}^i$ are these data samples and the graph edge weights $\mathcal{E}^i=\lbrace w_{k,l}^i\rbrace$, with  $w_{k,l}^i=K(\mathbf{x}_k^i,\mathbf{x}_l^i)$ as the elements of the adjacency matrix $\mathbf{W}^i$, indicate the neighborhood relations between samples and their measure of similarity using a kernel function $K(\cdot,\cdot)$.\par

The symmetric normalized graph Laplacian matrix for modality $i$ is defined as   $\mathbf{L}^i={\mathbf{D}^i}^{-1/2} (\mathbf{D}^i-\mathbf{W}^i){\mathbf{D}^i}^{-1/2}$, by discretizing the Laplace-Beltrami (LB) operator \citep{rosenberg1997}, where $\mathbf{D}^i=diag(\sum_{k\neq l} w_{k,l}^i)$ is the diagonal matrix of nodes degrees. For all $m$ Laplacian matrices $\lbrace{\mathbf{L}^i\in \mathbb{R}^{N_i\times N_i}}\rbrace_{i=1}^m$, there are unitary eigenspaces $\mathbf{U}^i$’s, such that $\mathbf{L}^i=\mathbf{U}^i \mathbf{\Lambda}^i {\mathbf{U}^i}^T$, where matrix $\mathbf{U}^i=[\mathbf{u}_1^i ... \mathbf{u}_{N_i}^i]$ contains the eigenvectors of $\mathbf{L}^i$ as its columns and diagonal matrix $\mathbf{\Lambda}^i=diag(\mathbf{\lambda}_1^i,…,\mathbf{\lambda}_{N_i}^i)$ contains their corresponding eigenvalues. \par

The spectral geometry of data in modality $i$ is indicated by the graph Laplacian spectrum defined as its eigenvalues, which describes the relationships between such a spectrum and the geometric structure of $\mathcal{M}^i$ \citep{rosenberg1997}. \par

Local descriptors (point signatures) are the feature vectors representing the local structure of the data graph, defined on the vertices to represent their local spectral geometry.

\subsection{Spectral graph wavelet signatures}
\label{3.3SpectralGraphWaveletSignatures}
Wavelet transform, as a powerful multiresolution analysis tool, can express a signal as a combination of several localized shifted and scaled versions of a simple function called a wavelet basis signal \citep{Front2009}. \par

Spectral graph wavelet transforms \citep{HAMMOND2011} are defined based on the eigenproblem of the graph Laplacian matrix, and can be applied as the scaling operations on a signal defined on the graph vertices. \par

As shown in \citep{Masoumi2017}, the spectral graph wavelet localized at vertex $r$ with scale parameter $\eta$ is given by:

\begin{equation}
\label{eq1}
\psi_{\eta,r}(t)=\sum_{l=1}^N g(\eta \lambda_l) u_l^*(r)u_l(t),
\end{equation}

\noindent where $t$ is a vertex of the graph, $N$ is the number of data samples, $\lambda_l$ is the $l$-th eigenvalue of the normalized graph Laplacian matrix and $\mathbf{u}_l$ is its associated eigenvector that its $t$-th element $u_l(t)$ is the value of the LB operator eigenfunction at vertex $t$, $*$ denotes complex conjugate operator (for real-valued eigenfunctions we have $u_l^*(t)=u_l(t))$, and $g:\mathbb{R}^+\rightarrow\mathbb{R}^+$ is the spectral graph wavelet generating kernel. The kernel $g$ should behave as a band-pass filter when satisfying $g(0)=0$ and $\lim_{x \to \infty} g(x)=0$ (we used the Mexican hat form as mentioned in section \ref{5.3.1FMBSDHyperParameterTuning}). The spectral graph wavelet coefficients of a given function $f$ are generated from its inner product with the spectral graph wavelets of equation (\ref{eq1}), as:

\begin{equation}
\label{eq2}
W_f(\eta,r)=\langle f,\psi_{\eta,r} \rangle=\sum_{l=1}^N g(\eta \lambda_l) \hat{f}(l) u_l(r),
\end{equation}

\noindent where $\hat{f}(l)$ is the value of graph Fourier transform of $f$ at eigenvalue $\lambda_l$ defined as follows:

\begin{equation}
\label{eq3}
 \hat{f}(l)=\langle \mathbf{u}_l,f \rangle=\sum_{i=1}^N u_l^*(i) f(i).
\end{equation}

\noindent Similarly, the scaling function at vertex $r$ is given by:

\begin{equation}
\label{eq4}
\zeta _r(t)=\sum_{l=1}^N h(\lambda_l) u_l^*(r)u_l(t),
\end{equation}

\noindent where function $h:\mathbb{R}^+\rightarrow\mathbb{R}$, which should satisfy $h(0)>0$ and $\lim_{x \to \infty} h(x)=0$, is used as a low-pass filter to better encode the low-frequency content of the function $f$ defined on the graph (we adopted the form given in \citep{HAMMOND2011}). The scaling coefficients of a given function $f$ are defined as inner product of this function with the scaling function of equation (\ref{eq4}) as follows:

\begin{equation}
\label{eq5}
S_f(r)=\langle f,\zeta_{r} \rangle=\sum_{l=1}^N h(\lambda_l) \hat{f}(l) u_l(r).
\end{equation}

To represent the localized structure around a graph vertex $r\in \mathcal{V}$, assume the unit impulse function $\delta_r$ centered at vertex $r$ as the function on the graph, i.e.  $f(t)=\delta_r (t)$ at each vertex $t \in \mathcal{V}$. Since according to equation (\ref{eq3}) we have $\hat{\delta}_r(l)=\sum_{i=1}^N u_l^*(i) \delta_r(i)=u_l^*(r)$, by substituting it in equations (\ref{eq2}) and (\ref{eq5}), the spectral graph wavelet coefficients $W_{\delta_r} (\eta,r)$ and scaling coefficients $S_{\delta_r}(r)$ are given as follows:

\begin{equation}
\label{eq6}
W_{\delta_r} (\eta,r)=\langle \delta_r,\psi_{\eta,r} \rangle=\sum_{l=1}^N g(\eta \lambda_l) u_l^2(r),
\end{equation}

\begin{equation}
\label{eq7}
S_{\delta_r} (r)=\langle \delta_r,\zeta_{r} \rangle=\sum_{l=1}^N h( \lambda_l) u_l^2(r).
\end{equation}

The spectral graph wavelet and scaling function coefficients for delta function are collected to form the \textit{spectral graph wavelet signature} (SGWS) as follow:

\begin{equation}
\label{eq8}
\mathcal{S}_r =\lbrace \mathbf{\upomega}_r^L \mid L=1,...,R \rbrace,
\end{equation}

\noindent where $R$ is the resolution parameter, and $\mathbf{w}_r^L$ is the graph signature vector at resolution level $L$, defined as:

\begin{equation}
\label{eq9}
\mathbf{\upomega}_r^L=\left[ W_{\delta_r} (\eta_1,r),...,W_{\delta_r} (\eta_L,r),S_{\delta_r} (r)\right],
\end{equation}

\noindent where $\eta_k$ is the wavelet scale whose resolution is determined by parameter $L$.

\subsection{Functional map}
\label{3.4FunctionalMap}
A functional map between two manifolds $\mathcal{M}^i$ and $\mathcal{M}^j$ representing two modalities $i$ and $j$ respectively, is defined as a generalized linear operator of classical pointwise mapping between data points of modalities $i$ and $j$ \citep{Ovsjanikov2012}.\par

By considering a correspondence or pointwise mapping $T:\mathcal{M}^j\rightarrow\mathcal{M}^i$ between data points on these manifolds, each point $\mathbf{p}$ on $\mathcal{M}^j$ can be transformed to the point $T(\mathbf{p})$ on $\mathcal{M}^i$.\par
Instead of using correspondences between data points, this mapping allows information transfer through manifolds across the two modalities. The information contains encoded geometry of the manifolds.\par

Using mapping $T$ from $\mathcal{M}^j$ to $\mathcal{M}^i$, each scaler function $f^i:\mathcal{M}^i\rightarrow \mathbb{R}$  can be transformed to the scaler function $f^j:\mathcal{M}^j\rightarrow \mathbb{R}$ simply via composition $f^j=f^i oT$, which means that $f^j (\mathbf{p})=f^i (T(\mathbf{p}))$ for any point $\mathbf{p}$ on $\mathcal{M}^j$.\par

A linear mapping between scalar function spaces $T_F:\mathbb{R}\rightarrow \mathbb{R}$ is defined as $f^j={T_F} (f^i)=f^i oT$ \citep{Ovsjanikov2012}. In a smilar manner, we propose an induced transformation that transforms the $\xi_i$-dimensional vector-valued functions $\mathbf{f}^i:\mathcal{M}^i\rightarrow \mathbb{R}^{\xi_i}$ defined on modality $i$ to the $\xi_j$-dimensional vector-valued functions $\mathbf{f}^j:\mathcal{M}^j\rightarrow \mathbb{R}^{\xi_j}$ defined on modality $j$ as $T_{VF}:\mathbb{R}^{\xi_i}\rightarrow \mathbb{R}^{\xi_j}$ defined as $\mathbf{f}^j=T_{VF} (\mathbf{f}^i)$. This is the functional representation of the original mapping $T:\mathcal{M}^j\rightarrow \mathcal{M}^i$ for the vector-valued signals on the manifolds. \par

The key idea of the functional map is a generalization of the notion of the map that is a correspondence between functions over one manifold to functions over another manifold, unlike the basic output of correspondence tools, which is a mapping from data points on one manifold to data points on another. \par

The matrix $\mathbf{C}$ can fully encode the original map $T_F$ such that for any function $f$ represented as a vector of coefficients $\mathbf{a}$, $T_F (\mathbf{a})=\mathbf{C}^T \mathbf{a}$ \citep{Ovsjanikov2012}. Similarly, we use a matrix representation of vector-valued signal $\mathbf{f}$ denoted by $\mathbf{A}$, to represent the functional mapping $T_{VF} (\mathbf{A})=\mathbf{C}^T \mathbf{A}$. This representation is used in next section to define the proposed objective function (c.f. equation (\ref{eq13})).

\section{Proposed methods}
\label{4ProposedMethods}

Two proposed methods are presented in this section. The first method is based on functional mapping between SGWS descriptors (FMBSD), which these descriptors represent the localities of each manifold. FMBSD method is a regularization-based framework that uses the functional mapping between SGWS descriptors for finding the pointwise correspondences, and is evaluated on the cross-modal information retrieval problems. The second method is manifold regularized multimodal classification based on pointwise correspondences (M$^2$CPC). M$^2$CPC applies the resulting pointwise correspondences from the FMBSD method in a new manifold regularized multimodal multi-class classification.

\subsection{Functional mapping between SGWS descriptors (FMBSD) for finding pointwise correspondences}
\label{4.1FMBSDForFindingPointwiseCorrespondences}

The purpose of this section is to find the pointwise correspondences between modalities related to the same phenomenon. These pointwise correspondences provide rich information about the relationships between different modalities in multimodal problems.\par
Our key idea in addressing this problem is to learn a mapping matrix between local descriptors of each pair of modalities while it is independent of prior knowledge about pointwise correspondences in the learning phase.\par
Inspired by this fundamental aspect of the functional map representation, local descriptor defined on each manifold can be approximated using a small number of bases functions \citep{Ovsjanikov2012}. As detailed in step 3, in this method, bases functions of each modality are computed based on the Laplace-Beltrami operator of its underlying data manifold.\par
The pointwise correspondences between two modalities can be obtained through the mapping of bases functions in one (source) modality to the bases functions in another (target) modality.\par
\begin{figure}[t!]
\centering
\includegraphics[width=\textwidth]{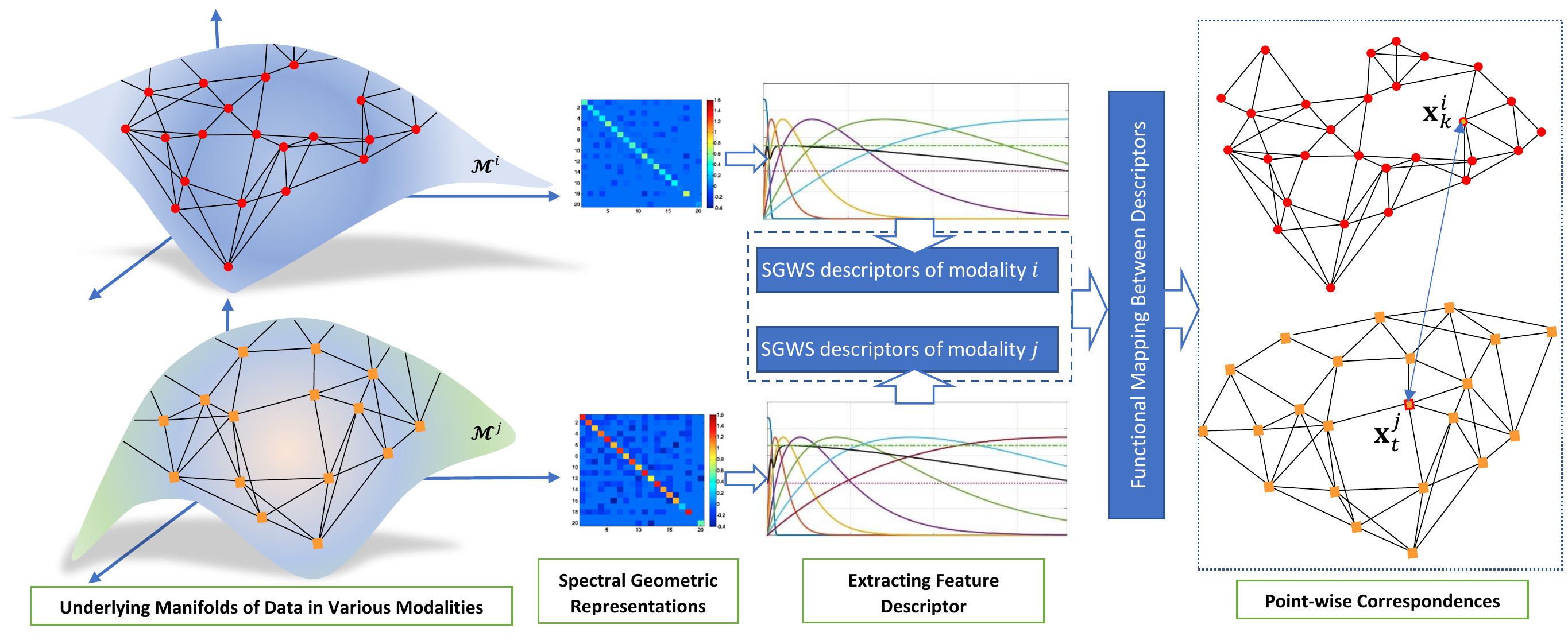}
\caption{The workflow of the proposed FMBSD method. For each manifold of data, represented by graph, the local SWGS descriptors at resolution $R$ is first extracted, then a functional mapping between these local descriptors is learned, which is used finally for pointwise correspondences }
\label{fig2}
\end{figure}

The workflow of our proposed FMBSD method is shown in Fig. \ref{fig2}. We present it step by step as follows.\\

\paragraph{\textbf{Step 1: Spectral geometric representations}.} In this step, first, the data neighborhood graphs for all modalities $\left\{\mathcal{G}^i\right\}_{i=1}^m$ representing their underlying data manifolds $\left\{\mathcal{M}^i\right\}_{i=1}^m$ are computed. Then, graph Laplacian matrices $\left\{\mathbf{L}^i\right\}_{i=1}^m$ are obtained as mentioned in section \ref{3.2DataSpectralGeometry}. \par
The eigen-decomposition of each Laplacian $\mathbf{L}^i$ is performed, and the top $k_i<N_i$ eigenvectors are stored as columns of the matrix $\mathbf{\Delta}^i=[\mathbf{u}_1^i ... \mathbf{u}_{k_i}^i]$. \\

\paragraph{\textbf{Step 2: Feature descriptors}.} One of the most important parts of the proposed approach is finding the functions that locally describe the data in each modality. In this paper, we use the spectral graph wavelet signature (SGWS) method to extract local descriptors of data in each modality, with the generating kernel $g$ and the scaling function $h$ used in \citep{Masoumi2017}. \par
As mentioned in section \ref{3.3SpectralGraphWaveletSignatures}, a data neighborhood graph $\mathcal{G}^i=(\mathcal{V}^i,\mathcal{E}^i)$ in $i$–th modality can be represented by the following $N_i\times q$ matrix of the SGWS descriptors at resolution $R$ as follows:

\begin{equation}
\label{eq10}
\mathbf{S}^i=\begin{bmatrix}W_{\delta_1}(\eta_1,1) & W_{\delta_1}(\eta_2,1)& \dotsm & W_{\delta_1}(\eta_R,1) & S_{\delta_1}(1)\\\vdots & \vdots & \ddots & \vdots & \vdots \\ W_{\delta_{N_i}}(\eta_1,{N_i}) & W_{\delta_{N_i}}(\eta_2,{N_i})& \dotsm & W_{\delta_{N_i}}(\eta_R,{N_i}) & S_{\delta_{N_i}}({N_i})\end{bmatrix},
\end{equation}

\noindent where $q=R+1$ is the number of columns and the $q$-dimensional feature vector in the $r$-th row of $\mathbf{S}^i$ is the local descriptor that encodes the local structure around the $r$-th vertex of $\mathcal{G}^i$. \par

By showing the above matrix as $\mathbf{S}^i=[\mathbf{s}_1^i ... \mathbf{s}_q^i ]\in \mathbb{R}^{N_i\times q}$, where $\mathbf{s}_l^i$ is the $l$-th column of $\mathbf{S}^i$ which indicates the $l$-th local descriptor function of all $N_i$ data samples. In this way, the data in modality $i$ is represented by $q$ local descriptor functions.\\

\paragraph{\textbf{Step 3: Spectral bases functions}.} As noted above, in comparison with a correspondence map between data samples, a functional map is a more general map between pairs of modalities, which is not limited to data samples. A functional map cannot be associated with any pointwise correspondence map because functional maps must preserve the set of values of each function.\par
To recover the pointwise correspondence representation by a functional map and make a functional map intuition practical, the size of the matrix that represents functional map must be moderate and ideally independent of the number of points on the manifold. To address this problem, a descriptor function defined on the underlying manifold in each modality is approximated using a sufficient number of bases functions.\par
Since the eigenvectors and eigenvalues of graph Laplacian can be considered as Fourier bases functions and frequencies, respectively, in this paper, a sufficient number of these bases functions, denoted by $k_i$, are used to approximate descriptor functions. For example, the $l$-th descriptor function $\mathbf{s}_l^i$ that is the $l$-th column of matrix $\mathbf{S}^i$ in (\ref{eq10}), can be approximated as:

\begin{equation}
\label{eq11}
\mathbf{s}_l^i\approx\sum_{t=1}^{k_i} a_{l,t}^i \mathbf{u}_t^i=\mathbf{\Delta}^i\mathbf{a}_l^i,
\end{equation}

\noindent where $\mathbf{a}_l^i=[a_{l,1}^i ... a_{l,k_i}^i]^T$ are Fourier coefficients, and $\mathbf{\Delta}^i=[\mathbf{u}_1^i ... \mathbf{u}_{k_i}^i]$ contains the most $k_i$ important eigenvectors $\mathbf{u}_t^i$, $1\leq t \leq k_i$ (corresponding to the least eigenvalues) of $\mathbf{L}^i$ as their columns. \par

According to the orthonormal property of eigenvectors, $\mathbf{a}_l^i={\mathbf{\Delta}^i}^T \mathbf{s}_l^i$ and $\mathbf{A}^i=[\mathbf{a}_1^i ... \mathbf{a}_q^i]$ collects the Fourier coefficients of all descriptor functions ($\mathbf{a}_l^i$'s are stored as columns of coefficients in the matrix $\mathbf{A}^i$).\\

\paragraph{\textbf{Step 4: Formulating the problem}.} Our purpose in this step is formulating the problem using manifold optimization. The objective is typically encoding information (e.g., geometry or appearance) that must be preserved by the unknown mapping. According to this objective function, we expect that geometric or appearance properties (descriptors) of manifolds be preserved by the map. \par
According to manifold optimization, two constraints should be satisfied simultaneously to control the between-modality and within-modality similarity relationships among data samples on two different modalities. Applying these constraints reduces the search space to find an accurate map that well represents the sample correspondences. \par
The optimal mapping matrix $\mathbf{C}^{i,j}$ with size $k_i\times k_j$ that maps Fourier coefficients of all descriptor functions from modality $i$ to modality $j$ is found by solving the following optimization problem:

\begin{equation}
\label{eq12}
\mathbf{C}_{opt}^{i,j}=\text{arg}\min_{\mathbf{C}^{i,j}}\mathcal{O}(\mathbf{C}^{i,j})+\Omega(\mathbf{C}^{i,j}).
\end{equation}

\noindent The first term in this problem is the objective function that composed of two components as follow:

\begin{equation}
\label{eq13}
\mathcal{O}(\mathbf{C}^{i,j})=\alpha \parallel {\mathbf{C}^{i,j}}^T\mathbf{A}^i -\mathbf{A}^j\parallel_F^2+\beta \sum_{k=1}^q \parallel \mathbf{\Phi}_k^i \mathbf{C}^{i,j}-\mathbf{C}^{i,j}\mathbf{\Phi}_k^j \parallel_F^2,
\end{equation}

\noindent where $\mathbf{\Phi}_k^i={\mathbf{\Delta}^i}^+ diag(\mathbf{s}_k^i) \mathbf{\Delta}^i$, which ${\mathbf{\Delta}^i}^+$ denotes the Moore-Penrose pseudoinverse of $\mathbf{\Delta}^i$. \par

Equation (\ref{eq13}) formulates a functional geometry preservation problem where encoded information of geometric structure in local descriptors, must be preserved by the unknown map $\mathbf{C}^{i,j}$. The first term in the equation (\ref{eq13}) leads to a linear system that computes Euclidean distance between mapped Fourier coefficients of modality $i$ to modality $j$ and Fourier coefficients of modality $j$.\par

As proved in \citep{Nogneng13124}, for obtaining a more accurate map by fewer independent descriptor functions, the second term, which encoded via commutativity of unknown map $\mathbf{C}^{i,j}$ is included in the equation (\ref{eq13}). Since obtaining a large set of independent descriptor functions is problematic, this term is used to avoid badly defined optimization problems, especially when the number of descriptor functions $q$ is smaller than the number of bases functions $k_j$.\par

The second term of the equation (\ref{eq12}) is the manifold regularization term, which is decomposed into two components:

\begin{equation}
\label{eq14}
\Omega(\mathbf{C}^{i,j})=\lambda_B \Omega_B(\mathbf{C}^{i,j})+\lambda_W \Omega_W(\mathbf{C}^{i,j}),
\end{equation}

\noindent where $\Omega_B(\cdot)$ and $\Omega_W(\cdot)$ indicate between-modality and within-modality similarity relationships, respectively, linearly combined by coefficients $\lambda_B$ and $\lambda_W$. Detailed discussions about these two functions are given below. \\

\paragraph{\textbf{Between-modality similarity relationship}} \mbox{}\par\nobreak
\noindent Although data in different modalities have different representations and are given in different feature spaces, there must be relationships between them since those are different aspects of the same content and share similar semantics. This is known as the between-modality similarity relationship. According to the between-modality similarity relationship between two modalities $i$ and $j$, the similarity of each pair of data after finding pointwise correspondences must be proportional to their prior similarity rate given in matrix $\mathbf{\Pi}^{i,j}$ with elements $\pi_{r,t}^{i,j}$ indicating the prior similarity between $r$-th sample in modality $i$ and $t$-th sample in modality $j$, computed based on the equation (\ref{eq18}).\par
If the above mentioned two samples are similar indicated by high value of $\pi_{r,t}^{i,j}$, then the mapping matrix $\mathbf{C}^{i,j}$ should be found in such a way that the low-dimensional representation of the $r$-th sample in modality $j$ shown by $\mathbf{\upgamma}_r^{i,j}$ becomes close to the mapped low-dimensional representation of the $t$-th sample in modality $i$ shown by $\mathbf{\updelta}_t^j$. Accordingly, the between modality similarity term defined as follows is minimized:

\begin{equation}
\label{eq15}
\Omega_B(\mathbf{C}^{i,j})=\sum_{r=1}^{N_i}\sum_{t=1}^{N_j}\pi_{r,t}^{i,j} \parallel \upgamma_r^{i,j}-\updelta_t^j\parallel^2,
\end{equation}

\noindent where $\mathbf{\upgamma}_r^{i,j}$ ($1\leq r\leq N_i$) with dimension $1\times k_j$ is the $r$-th row of $\mathbf{\Gamma}^{i,j}=\mathbf{\Delta}^i \mathbf{C}^{i,j}$ and $\mathbf{\updelta}_t^j$ ($1\leq t \leq N_j$) with dimension $1\times k_j$ is the $t$-th row of $\Delta^j$. As it will be proved in Appendix \ref{appA}, equation (\ref{eq15}) is equivalent to the following equation written in the matrix form:

\begin{equation}
\label{eq16}
  \Omega_B(\mathbf{C}^{i,j})=tr\left({\mathbf{\Gamma}^{i,j}}^T \mathbf{D}^i \mathbf{\Gamma}^{i,j}-2{\mathbf{\Gamma}^{i,j}}^T\mathbf{\Pi}^{i,j}\mathbf{\Delta}^{j}+{\mathbf{\Delta}^{j}}^T\tilde{\mathbf{D}}^{j}\mathbf{\Delta}^{j}\right),
\end{equation}

\noindent where $\mathbf{D}^i=diag(d_1^{i,j},...,d_{N_i}^{i,j})$ is diagonal matrix of size $N_i\times N_i$ with $d_r^{i,j}=\sum_{t=1}^{N_j} \pi_{r,t}^{i,j}$, and similarly $\tilde{\mathbf{D}}^j=diag(\tilde{d}_1^{i,j},...,\tilde{d}_{N_j}^{i,j})$ with $\tilde{d}_r^{i,j}=\sum_{r=1}^{N_i} \pi_{r,t}^{i,j}$.\par

All entries of similarity matrix $\mathbf{\Pi}^{i,j}$ are computed based on our following proposed similarity measure. The dissimilarity between $r$-th data sample in modality $i$ and $t$-th data point in modality $j$ can be calculated given spectral graph wavelet signatures of two modalities at resolution level $L$, as:

\begin{equation}
\label{eq17}
d_{SWGS}(\mathbf{x}_r^i,\mathbf{x}_t^j)=\sum_{k=1}^q\left(s_k^i(r)-s_k^j(t)\right)^2,
\end{equation}

\noindent where $s_k^i(r)$ is the $r$-th element of the $k$-th local descriptor $\mathbf{s}_k^i$ in modality $i$, with $1\leq r \leq N_i$.\par

\noindent The similarity between two points $\mathbf{x}_r^i$ and $\mathbf{x}_t^j$ are then given by:

\begin{equation}
\label{eq18}
\pi_{r,t}^{i,j}=exp\left[\frac{-d_{SWGS}^2(\mathbf{x}_r^i,\mathbf{x}_t^j) }{ 2\sigma_{SWGS}^2}\right],
\end{equation}

\noindent where $\sigma_{SWGS}$ is the computed sample variance of $d_{SWGS}$ given all samples.

\paragraph{\textbf{Within-modality similarity relationship}} \mbox{}\par\nobreak
\noindent According to the within-modality similarity relationship, the similarity among data points within every single modality is expected to be preserved through the mapping, i.e., the data points with the neighborhood relationship should be close to each other after mapping. To preserve the local structural information within each single modality, the within-modality similarity term is defined as follows:

\begin{equation}
\label{eq19}
\Omega_W(\mathbf{C}^{i,j})=\sum_{r,t}w_{r,t}^i \parallel \mathbf{\upgamma}_r^{i,j}-\mathbf{\upgamma}_t^{i,j}\parallel^2=tr\left({\mathbf{\Gamma}^{i,j}}^T\mathbf{L}^i \mathbf{\Gamma}^{i,j}\right),
\end{equation}

\noindent By substituting $\mathbf{\Gamma}^{i,j}=\mathbf{\Delta}^i \mathbf{C}^{i,j}$ in (\ref{eq16}) and (\ref{eq19}), equation (\ref{eq14}) is as follows:

\begin{equation}
\label{eq20}
\begin{aligned}
\Omega(\mathbf{C}^{i,j})=&\lambda_B tr\left({\mathbf{C}^{i,j}}^T {\mathbf{\Delta}^{i}}^T \mathbf{D}^{i}\mathbf{\Delta}^{i}\mathbf{C}^{i,j}-2{\mathbf{C}^{i,j}}^T {\mathbf{\Delta}^{i}}^T \mathbf{\Pi}^{i,j}\mathbf{\Delta}^{j}+{\mathbf{\Delta}^{j}}^T \tilde{\mathbf{D}}^{j}\mathbf{\Delta}^{j}\right)+\\
&\lambda_W tr\left({\mathbf{C}^{i,j}}^T{\mathbf{\Delta}^{i}}^T\mathbf{L}^{i} \mathbf{\Delta}^{i}\mathbf{C}^{i,j}\right),
\end{aligned}
\end{equation}

\noindent where the similarity $w_{r,t}^i$ between each two samples $\mathbf{x}_r^i$ and $\mathbf{x}_t^i$ in modality $i$ is define as:

\begin{equation}
\label{eq21}
w_{r,t}^i=\begin{cases}exp\left( \frac{-z_{r,t}^i}{2\sigma^2}\right) & { if \;  \mathbf{x}_r^i\in N_k(\mathbf{x}_t^i) \; or \;  \mathbf{x}_t^i\in N_k(\mathbf{x}_r^i})\\0 &otherwise\end{cases},
\end{equation}

\noindent where $z_{r,t}^i=\parallel \mathbf{x}_t^i-\mathbf{x}_r^i \parallel^2$ is the squared Euclidean distance between $\mathbf{x}_t^i$ and $\mathbf{x}_r^i$, $N_k (\mathbf{x}_t^i)$ denotes the set of $k$ nearest neighbors of $\mathbf{x}_t^i$, and $\sigma$ is an extent parameter, which set to the mean of distances between each sample and its nearest samples.\par

\noindent By substituting equations (\ref{eq16}) and (\ref{eq20}) in equation (\ref{eq14}), problem (\ref{eq12}) is rewritten as:

\begin{equation}
\label{eq22}
\begin{aligned}
\mathbf{C}_{opt}^{i,j}=&\text{arg}\min_{\mathbf{C}^{i,j}} \;\alpha\parallel {\mathbf{C}^{i,j}}^T\mathbf{A}^i -\mathbf{A}^j\parallel_F^2+\beta \sum_{k=1}^q \parallel \mathbf{\Phi}_k^i \mathbf{C}^{i,j}-\mathbf{C}^{i,j}\mathbf{\Phi}_k^j \parallel_F^2+\\
&\lambda_B tr\left({\mathbf{C}^{i,j}}^T {\mathbf{\Delta}^{i}}^T \mathbf{D}^{i}\mathbf{\Delta}^{i}\mathbf{C}^{i,j}-2{\mathbf{C}^{i,j}}^T {\mathbf{\Delta}^{i}}^T \mathbf{\Pi}^{i,j}\mathbf{\Delta}^{j}+{\mathbf{\Delta}^{j}}^T \tilde{\mathbf{D}}^{j}\mathbf{\Delta}^{j}\right)+\\
&\lambda_Wtr\left({\mathbf{C}^{i,j}}^T{\mathbf{\Delta}^{i}}^T\mathbf{L}^{i} \mathbf{\Delta}^{i}\mathbf{C}^{i,j}\right).
\end{aligned}
\end{equation}

\paragraph{\textbf{Step 5: Optimization for finding the map}.} In this step, we find the solution $\mathbf{C}_{opt}^{i,j}$ for problem (\ref{eq22}) using an iterative continuous optimization algorithm that solves the minimization problem. Since this problem is convex, many optimization methods, containing the gradient-based ones, can ensure to achieve the global optimum.\par
Gradient-based methods require access to the derivatives of the objective function $\mathcal{O}(\mathbf{C}^{i,j})$ and manifold regularization $\Omega(\mathbf{C}^{i,j})$. As it will be proved in Appendix \ref{appB}, derivatives of the objective function (\ref{eq13}) and manifold regularization (\ref{eq20}) with respect to $\mathbf{C}^{i,j}$ are obtained as follows:

\begin{equation}
\label{eq23}
\begin{aligned}
&\triangledown_{\mathbf{C}^{i,j}}\mathcal{O}(\mathbf{C}^{i,j})=2\alpha{\mathbf{A}^{i}}\left( {\mathbf{C}^{i,j}}^T\mathbf{A}^i -\mathbf{A}^j\right)^T\\
&+2\beta \left( \sum_{k=1}^q {\mathbf{\Phi}_k^i}^T  \left(\mathbf{\Phi}_k^i\mathbf{C}^{i,j}-\mathbf{C}^{i,j}\mathbf{\Phi}_k^j \right)-\sum_{k=1}^q   \left(\mathbf{\Phi}_k^i\mathbf{C}^{i,j}-\mathbf{C}^{i,j}\mathbf{\Phi}_k^j \right) {\mathbf{\Phi}_k^j}^T\right),
\end{aligned}
\end{equation}

\begin{equation}
\label{eq24}
\begin{aligned}
\triangledown_{\mathbf{C}^{i,j}}\Omega(\mathbf{C}^{i,j})=
2\lambda_B \left({\mathbf{\Delta}^{i}}^T \mathbf{D}^{i}\mathbf{\Delta}^{i}\mathbf{C}^{i,j}-{\mathbf{\Delta}^{i}}^T \mathbf{\Pi}^{i,j}\mathbf{\Delta}^{j}\right)+2
\lambda_W {\mathbf{\Delta}^{i}}^T\mathbf{L}^{i} \mathbf{\Delta}^{i}\mathbf{C}^{i,j}.
\end{aligned}
\end{equation}

\noindent Using these two derivatives, we calculate the solution of problem (\ref{eq22}) by applying the standard constrained optimization tools such as \textit{fmincon} of MATLAB.

\paragraph{\textbf{Step 6: Pointwise correspondences}.} In this formulation, two matrices $\mathbf{\Delta}^i$ and $\mathbf{\Delta}^j$ are interpreted as the spectral embeddings of the two modalities $i$ and $j$, respectively, and the functional map $\mathbf{C}^{i,j}$ that is represented by the spectral coefficients between modalities $i$ and $j$ is applied to align these two embeddings.\par

The $t$-th sample in modality $j$ is represented by $t$-th row of $\mathbf{\Delta}^j$ as its dimensionality reduced version. By adopting the simple nearest-neighbor method used in \citep{Ovsjanikov2012}, if the $k$-th row in $\mathbf{\Delta}^i \mathbf{C}^{i,j}$ has the shortest Euclidean distance with $t$-th row of $\mathbf{\Delta}^j$, $k$-th sample in modality $i$ is considered as the corresponding with $t$-th sample in modality $j$, and $\mathbf{\uprho}^{i,j} (k)=t$, which $\mathbf{\uprho}^{i,j}$ is a pointwise correspondence vector with $N_i$ elements between modalities $i$ and $j$.\par

In order to represent pointwise correspondences between each two modalities $i$ and $j$ ($i<j$), we define the $N_i\times N_j$  pointwise correspondences matrix $\mathbf{P}^{i,j}$ as follows:

\begin{equation}
\label{eq25}
\mathbf{P}^{i,j}=\left[ \mathbf{p}_1^{i,j} \; \mathbf{p}_2^{i,j} \; ... \;  \mathbf{p}_{N_j}^{i,j}\right],
\end{equation}

\noindent where $\mathbf{p}_k^{i,j}$ contains $1$ at index $t$ if $\mathbf{\uprho}^{i,j} (k)=t$, and $0$ otherwise. \par

Using the mapping matrices between modality $i$ and all other modalities, and simply mapping $\mathbf{\Delta}^i$ to other modalities (target modalities), pointwise correspondences between data samples in modality $i$ and those in other modalities can be obtained. All mentioned steps are summarized in Algorithm \ref{alg1}.

\begin{algorithm}[h]
  \caption{Functional Mapping between SGWS descriptors for finding pointwise correspondences (FMBSD)}
  \label{alg1}
  \begin{algorithmic}[h]
    \State \textbf{Input:}
    \begin{itemize} [label=$-$]
        \item $\lbrace \mathbf{X}^i \rbrace_{i=1}^m$: Training data from $m$ modalities ($m$: Number of modalities).
    \end{itemize}
    \State \textbf{Hyper-parameters:}
     \begin{itemize} [label=$-$]
        \item $g(\cdot)$: Kernel generating function
        \item $h(\cdot)$: Scaling function
        \item  $R$: Resolution parameter
        \item $k_i$: Number of eigenvectors selected (lower dimension) in modality $i$ ($1\leq i\leq m$)
    	\item $k$: Number of nearest neighbors
        \item $\alpha$, $\beta$, $\lambda_W$, and $\lambda_B$: Regularization parameters
    \end{itemize}
    \State \textbf{Procedure:}\\
    \textbf{1. For} each modality $i$ ($1\leq i \leq m$)
    \begin{enumerate}[label=\roman*.]
         \item Compute the Eigen-problem on Laplacian matrix $\mathbf{L}^i$ and store the top $k_i\ll N_i$ respected basis functions (eigenvectors) as columns of matrix  $\mathbf{\Delta}^i$.
         \item Compute a set of $q$ spectral graph wavelet signatures as local descriptors on manifold $\mathcal{M}^i$ and store them as columns of matrix $\mathbf{S}^i=[\mathbf{s}_1^i ... \mathbf{s}_q^i]\in \mathbb{R}^{N_i\times q}$ according to equation (\ref{eq10}).
         \item Approximate descriptors function by $k_i$ bases functions using equation (\ref{eq11}) and extract Fourier coefficients $\mathbf{A}^i$.
    \end{enumerate}
    \textbf{end for}\\
    \textbf{2. For} each pair of modalities $i$ and $j$ ($1\leq i,j \leq m$)
    \begin{enumerate}[label=\roman*.]
        \State 	\indent Solve the problem given in equation (\ref{eq22}) to find $\mathbf{C}_{opt}^{i,j}$ by using the gradients given in (\ref{eq23}) and (\ref{eq24}).
     \end{enumerate}
    \textbf{end for}
    \State 3. Find pointwise correspondences matrix $\mathbf{P}^{i,j}$ between each pair of modalities $i$ and ($1\leq i,j \leq m$), $i\neq j$ according to Step 6 mentioned in the text.
    \State \textbf{End}
    \State \textbf{Output:}
    \begin{itemize} [label=$-$]
    \item Pointwise correspondences matrices $\mathbf{P}^{i,j}$ ($1\leq i,j \leq m$).
    \end{itemize}

  \end{algorithmic}
\end{algorithm}

\subsubsection{Time complexity analysis}
\label{4.1.1TimeComplexityAnalysis}
According to equations (\ref{eq22}) and (\ref{eq23}), having $q$ descriptor functions and $k_i$ and $k_j$ bases functions in modality $i$ and $j$, respectively, computing the objective function and its gradients in each iteration of the optimization algorithm (for finding functional map step) have the complexity of $O(q k_i k_j)$.\par

Applying the $k-d$ tree search method \citep{Bentley1975} for finding the pointwise correspondences in Step 6, gives a significant efficiency improvement in practice and makes its complexity of $O \left( N_i \log N_i + N_j \log N_j \right)$. Since in this implementation, Laplacian eigenbases, SGWS descriptors, and their approximations for all modalities can be computed offline, total time complexity of Algorithm 1 for finding matrix $\mathbf{P}^{i,j}$ depends on its two steps 2 and 3 and is of order $O \left( q k_i k_j ,+ N_i \log N_i + N_j \log N_j \right)$.

\subsubsection{Connection with cross-modal retrieval problems}
\label{4.1.2ConnectionWithCross-modalRetrievalProblems}

Data samples on two different modalities are called relevant if they are semantically similar to each other. The purpose of cross-modal retrieval is finding more relevant data samples in each modality $j$ with the data samples in another modality $i$ ($i\neq j,1\leq i,j \leq m$). \par
Finding pointwise correspondences is a special case of the above mentioned cross-modal retrieval. The cross-modal retrieval in more general form can be achieved by extending Step 6 to find the $k$ samples in target modality that are the most similar ones to a query sample in source modality.

\subsection{Manifold regularized multimodal classification based on pointwise correspondences (M$^2$CPC)}
\label{4.2M2CPC}
The second proposed method introduces a framework for general multimodal manifold analysis based on pointwise correspondences between modalities described in section \ref{4.1FMBSDForFindingPointwiseCorrespondences}. Manifold regularized classification framework \citep{Mikhail2006} is originally designed for binary classification on single modality. Our approach extends this framework to vector-valued manifold regularization for multimodal multi-class data classification with properties mentioned in section \ref{1.intro}. \par

Our framework considers multi-class classification problem in the case that output (hypothesis) spaces are vector-valued reproducing kernel Hilbert spaces (RKHS) and data arise from heterogeneous modalities without any given explicit correspondences information (unlike multi-view problems). In our framework, 1) the target function used for label approximation is a vector in a Hilbert space of functions \citep{Carmeli2006} and heterogeneous data in different modalities are mapped into high dimensional spaces (vector-valued RKHS) by kernel transformation, and 2) pointwise correspondences used in the manifold regularization framework is determined based on our first proposed method in the previous section.\par

Following the definition of data in section \ref{3.1ProblemFormulation}, multimodal data set $\left\{\mathbf{X}^i\right\}_{i=1}^m$ is divided into two labeled and unlabeled data sets $\mathbf{Z}_l=\left\{\mathbf{z}_l^1,...,\mathbf{z}_l^m\right\}$ and $\mathbf{Z}_u=\left\{ \mathbf{z}_u^1,...,\mathbf{z}_u^m \right\}$, respectively, which $\mathbf{Z}=\mathbf{Z}_l\cup \mathbf{Z}_u$, $\mathbf{z}_l^i=\left\{(\mathbf{x}_j^i,\mathbf{y}_j^i)\right\}_{j=1}^{l_i}$, $\mathbf{x}_j^i\in \mathbb{R}^{d_i}$ is the $j$’th sample in modality $i$, $\mathbf{y}_j^i$  is the corresponding label of data sample $\mathbf{x}_j^i$, and $\mathbf{z}_u^i=\left\{ \mathbf{x}_j^i\right\}_{j=l_i+1}^{N_i}$.\par

Consider $\mathbf{y}_j^i\in \mathcal{Y}$ as the label vector of the $j$-th data sample in modality $i$. For a $c$-class classification problem that $\mathcal{Y}=\mathbb{R}^c$, the vector $\mathbf{y}^i=\lbrack {\mathbf{y}_1^i}^T ...{\mathbf{y}_{N_i}^i}^T\rbrack^T$ has $cN^i$ elements, where if $\mathbf{x}_j^i$ belongs to $k$-th class, then $\mathbf{y}_j^i=\lbrack -1,...,-1,1,-1,...-1\rbrack^T$ with $1$ in the $k$-th location and $-1$ in all other locations. Note that for the unlabeled data sample $\mathbf{x}_j^i$, the vector $\mathbf{y}_j^i$ contains zero in all $c$ locations.\par

According to vector-valued RKHS \citep{Minh2011,Minh2016}, consider $\mathcal{X}$ as input space, and $\mathcal{Y}$ as a separable Hilbert space denoting the output space with the inner product $\langle \cdot,\cdot\rangle_\mathcal{Y}$. Then, a $\mathcal{Y}$-valued RKHS $\mathcal{H}_K$ that is associated with an operator–valued positive definite kernel $K:\mathcal{X}\times \mathcal{X} \rightarrow \mathcal{L}(\mathcal{Y})$ measures a pairwise distance among data samples in each modality, where $\mathcal{L}(\mathcal{Y})$ is Banach space of bounded linear operator on $\mathcal{Y}$.

\subsubsection{Formulation of the proposed framework}
\label{4.2.1FormulationOfTheProposedFramework}

The purpose is approximating the label of data samples in modality $i$ by optimizing the target function $\mathbf{f}^i$ ($1\leq i\leq m$), which is a vector belonging to a vector-valued RKHS $\mathcal{H}_K$. \par

Consider $\mathbf{f}$ as a vector of target functions of all $m$ modalities as:

\begin{equation}
\label{eq26}
\mathbf{f}=\lbrack {\mathbf{f}^1}^T \; \dotsm \; {\mathbf{f}^m}^T   \rbrack^T \in \mathcal{Y}^N,
\end{equation}

\noindent where $N=\sum_{i=1}^m N_i$  and target function in each modality be a vector with the following structure:

\begin{equation}
\label{eq27}
\mathbf{f}^i=\lbrack {\mathbf{f}^i(\mathbf{x}_1^i)}^T \; \dotsm \; {\mathbf{f}^i(\mathbf{x}_{N_i}^i)}^T   \rbrack^T \in \mathcal{Y}^{N_i},
\end{equation}

\noindent where $\mathbf{f}^i(\mathbf{x}_j^i)\in \mathcal{Y}$ represents the target function of data sample $\mathbf{x}_j^i$ in modality $i$. \par

The general vector-valued manifold regularization framework for multimodal classification can be formulated as the following minimization problem:

\begin{equation}
\label{eq28}
\mathbf{f}_{opt}=\text{arg}\min_{\mathbf{f}\in\mathcal{H}_k} \frac{1}{l}\sum_{i=1}^m \sum_{j=1}^{l_i} V\left(\mathbf{y}_j^i, \mathbf{f}^i (\mathbf{x}_j^i) \right)+\gamma_A \parallel\mathbf{f}\parallel_{\mathcal{H}_k}^2+\gamma_W\langle \mathbf{f}, \mathbf{M}_W  \mathbf{f}\rangle_{\mathcal{Y}^N} +\gamma_B\langle \mathbf{f}, \mathbf{M}_B  \mathbf{f}\rangle_{\mathcal{Y}^N},
\end{equation}

\noindent where $V:\mathcal{Y}\times \mathcal{Y}\rightarrow \mathbb{R}$ is a convex loss function, $l=\sum_{i=1}^m l_i$, $\gamma_A$ is a regularization parameter that adjusts the smoothness of resulting target function in the ambient space (discussed in section 4.2.1.b), and $\gamma_w$ and $\gamma_B$ are regularization parameters that adjust a trade-off between simpler form and better fit to the intrinsic geometry of the the target function.\par

In problem (\ref{eq28}), two matrices $\mathbf{M}_W,\mathbf{M}_B:\mathcal{Y}^N\rightarrow \mathcal{Y}^N$ are the symmetric positive operators.\par

Let $\mathbf{M}:\mathcal{Y}^N\rightarrow \mathcal{Y}^N$ be a symmetric positive operator, that is $\langle \mathbf{y},\mathbf{My}\rangle_{\mathcal{Y}^N}\ge 0$ for all $\mathbf{y}\in \mathcal{Y}^N$. The operator $\mathbf{M}$ can be expressed as block matrix $\mathbf{M}=\left(\mathbf{M}^{i,j}\right)_{m\times m}$ such that:

\begin{equation}
\label{eq29}
\langle \mathbf{f}, \mathbf{M} \mathbf{f}\rangle_{\mathcal{Y}^N} =\sum_{i,j=1}^m \langle \mathbf{f}^{i}, \mathbf{M}^{i,j} \mathbf{f}^j\rangle_{\mathcal{Y}^{N_i}},
\end{equation}

\noindent where $\mathbf{M}^{i,j}=\left( \mathbf{M}_{t,r}^{i,j} \right)_{N_i\times N_j}$ is a linear operator, and since $dim(\mathcal{Y})=c$, each its element is a $c\times c$ matrix $\mathbf{M}_{t,r}^{i,j}$ such that:

\begin{equation}
\label{eq30}
\langle \mathbf{f}^{i}, \mathbf{M}^{i,j} \mathbf{f}^j\rangle_{\mathcal{Y}^{N_i}}=\sum_{t=1}^{N_i} \sum_{r=1}^{N_j}\langle \mathbf{f}^{i} (\mathbf{x}_t^i), \mathbf{M}_{t,r}^{i,j} \mathbf{f}^j  (\mathbf{x}_r^j)\rangle_{\mathcal{Y}}.
\end{equation}

In the following, we explain each part of the problem (\ref{eq28}) in detail, as well as reformulating for optimizing it.

\paragraph{\textbf{a) Loss function}} \mbox{}\par\nobreak
\noindent The first term in problem (\ref{eq28}) is the loss function, which measures the error between the target function $\mathbf{f}^i(\mathbf{x}_j^i)$ with the desired output $\mathbf{y}_i^j,1\leq i \leq l$. By considering $V(\cdot,\cdot)$ as a square loss function and $\mathbf{y}=\lbrack {\mathbf{y}^1}^T ... {\mathbf{y}^m}^T\rbrack^T$ as a vector of labels of data in all modalities, where $\mathbf{y}^i=\lbrack {\mathbf{y}_1^i}^T,...,{\mathbf{y}_{l_i}^i}^T,{\mathbf{y}_{l_i+1}^i}^T,...,{\mathbf{y}_{l_i+u_i}^i}^T \rbrack^T$ is a column vector of labels in modality $i$ with $cN_i$ elements, the loss function can be written as follows:

\begin{equation}
\label{eq31}
 \frac{1}{l}\sum_{i=1}^m \sum_{j=1}^{l_i} \parallel\mathbf{y}_j^i- \mathbf{f}^i (\mathbf{x}_j^i) \parallel_{\mathcal{Y}}^2= \frac{1}{l}\sum_{i=1}^m \parallel\mathbf{y}^i- \mathbf{f}^i  \parallel_{\mathcal{Y}}^2=\frac{1}{l} \parallel\mathbf{y}- \mathbf{Jf}  \parallel_{\mathcal{Y}}^2,
\end{equation}

\noindent where $\mathbf{J}:\mathbb{R}^{Nc\times Nc}$ is a block diagonal matrix as follows:

\begin{align*}
\mathbf{J} =\begin{bmatrix}\mathbf{J}^1 & \dotsm & 0\\\vdots & \ddots& \vdots \\ 0 & \dotsm & \mathbf{J}^m \end{bmatrix},
\end{align*}

\noindent where $\mathbf{J}^i$ ($1\leq i\leq m$), is a block diagonal matrix with $N_i$ diagonal blocks, which the first $l_i$ blocks are $c\times c$ identity matrices $\mathbf{I}_c$ and all entries of the rest diagonal block are $0$ as follows:

\begin{align*}
\mathbf{J}^i =\begin{bmatrix}\mathbf{I}_c & \dotsm & 0 & 0 & \dotsm & 0 \\ \vdots & \ddots& \vdots & \vdots & \ddots& \vdots \\ 0 & \dotsm & \mathbf{I}_c & 0 & \dotsm & 0 \\ 0 & \dotsm & 0 & 0 & \dotsm & 0\\ \vdots & \ddots& \vdots & \vdots & \ddots& \vdots\\0 & \dotsm & 0 & 0 & \dotsm & 0 \end{bmatrix}.
\end{align*}

\paragraph{\textbf{b) Ambient regularization}} \mbox{}\par\nobreak
\noindent The learning problem (\ref{eq28}) attempts to obtain a target function $\mathbf{f}$ among a RKHS hypothesis space $\mathcal{H}_K$ of functions. The ambient regularization term $\parallel \mathbf{f} \parallel_{\mathcal{H}_K}^2$ represents the complexity of the target function in the RKHS hypothesis space $\mathcal{H}_K$. \par
This term ensures that the solution is smooth with respect to ambient space by representing the complexity of target function $\mathbf{f}$ in the hypothesis space and penalizing its RKHS norm to impose smoothness conditions.

\paragraph{\textbf{c) Between modality regularization}} \mbox{}\par\nobreak
\noindent The third term in problem (\ref{eq28}) reflects the intrinsic structure of data between various modalities and measures the consistency of target function across them. For simplicity, we suppose $R_{\tilde{B}}(\mathbf{\tilde{f}})$ as between modality regularization in real-valued target function $\mathbf{\tilde{f}}=\lbrack \mathbf{\tilde{f}}^{1^T} \;...\; \mathbf{\tilde{f}}^{m^T} \rbrack^T \in \mathbb{R}^N$ for binary classification as follows:

\begin{equation}
\label{eq32}
R_{\tilde{B}}(\mathbf{\tilde{f}})=\sum_{ j,k=1,\; j<k }^{m} \sum_{i=1}^{N_j} \left( \tilde{f}^{j} (\mathbf{x}_i^j)-\mathbf{\tilde{f}}^{k} \left(\mathbf{x}_{\mathbf{\uprho}^{j,k}(i)}^k\right)\right)^2,
\end{equation}

\noindent where $\tilde{\mathbf{f}}^i=\lbrack\tilde{f}^i(\mathbf{x}_1^i) \; ... \; \tilde{f}^i(\mathbf{x}_{N_i}^i)\rbrack^T \in \mathbb{R}^{N_i}$ and $\mathbf{\uprho}^{j,k}$ is the pointwise correspondence vector, introduced in section \ref{4.1FMBSDForFindingPointwiseCorrespondences}. Using pointwise correspondences matrix $\mathbf{P}^{j,k}$ between each two modalities $j$ and $k$ in equation (\ref{eq25}), equation (\ref{eq32}) can be written as:

\begin{equation}
\label{eq33}
R_{\tilde{B}}(\mathbf{\tilde{f}})=\sum_{ j,k=1,\; j<k }^{m} \sum_{i=1}^{N_j} \left( \tilde{f}^{j} (\mathbf{x}_i^j)-{\mathbf{\tilde{f}}^{k^T}} \mathbf{p}_i^{j,k} \right)^2.
\end{equation}

\noindent We define the $N\times N$ block matrix $\mathbf{M}_p$, each block $\mathbf{M}_p^{j,k}$ is a $N_j\times N_k$ matrix, as follows:

\begin{align*}
\mathbf{M}_P =\begin{bmatrix}\mathbf{A}_1 & -{\mathbf{P}^{1,2}}^T & \dotsm & -{\mathbf{P}^{1,m-1}}^T & -{\mathbf{P}^{1,m}}^T  \\ -\mathbf{P}^{1,2} & \mathbf{A}_2 & \dotsm & -{\mathbf{P}^{2,m-1}}^T & -{\mathbf{P}^{2,m}}^T \\ \vdots & \vdots & \ddots & \vdots & \vdots  \\ -\mathbf{P}^{1,m-1} & -\mathbf{P}^{2,m-1} & \dotsm & \mathbf{A}_{m-1} & -{\mathbf{P}^{m-1,m}}^T \\ -\mathbf{P}^{1,m} & -\mathbf{P}^{2,m} & \vdots & -\mathbf{P}^{m-1,m} & \mathbf{A}_{m} \end{bmatrix},
\end{align*}

\noindent where $\mathbf{A}_i=(m-1)\mathbf{I}_{N_i}$ is a $N_i\times N_i$ diagonal matrix.\par

By considering matrix $\mathbf{M}_P$, between modality regularization in equation (\ref{eq33}) can be written as follows:

\begin{equation}
\label{eq34}
R_{\tilde{B}}(\tilde{\mathbf{f}})=\langle\mathbf{\tilde{f}},\mathbf{M}_P\mathbf{\tilde{f}}\rangle_{\mathbb{R}^N}=\mathbf{\tilde{f}}^T\mathbf{M}_P\mathbf{\tilde{f}}=\sum_{ j,k=1,\; j<k }^{m} \sum_{i=1}^{N_j} \left({\tilde{f}}^j (\mathbf{x}_i^j)-{\mathbf{\tilde{f}}^{k^T}} \mathbf{p}_i^{j,k} \right)^2.
\end{equation}

By extending equation (\ref{eq34}) to the vector-valued target function $\mathbf{f}\in \mathcal{Y}^N$ for $c$-class classification, $c\ge 2$, between modality regularization is defined as follows:

\begin{equation}
\label{eq35}
R_{B}(\mathbf{f})=\langle\mathbf{f},\mathbf{M}_B\mathbf{f}\rangle_{\mathcal{Y}^N}=\mathbf{f}^T\mathbf{M}_B\mathbf{f}=\sum_{ j,k=1,\; j<k }^{m} \sum_{i=1}^{N_j} \parallel{\mathbf{f}}^j (\mathbf{x}_i^j)-(\mathbf{p}_i^{j,k}\otimes \mathbf{I}_c){\mathbf{f}^{k}}  \parallel_{\mathcal{Y}}^2,
\end{equation}

\noindent where $\mathcal{Y}=\mathbb{R}^c$, $\mathbf{M}_B=\mathbf{M}_p\otimes \mathbf{I}_c$, $\mathbf{I}_c$ is the $c\times c$ identity matrix, and $\otimes$ is the Kronecker product.

\paragraph{\textbf{d) Within  modality regularization}} \mbox{}\par\nobreak
\noindent This term tries to preserve the intrinsic structure of data in each modality by measuring the smoothness of the target function in each modality and controlling its complexity with respect to its intrinsic geometry. \par
We suppose $R_{\tilde{W}}(\mathbf{\tilde{f}}^k)$ as within modality regularization in real-valued target function $\mathbf{\tilde{f}}^k\in \mathbb{R}^{N_k}$ as follows:

\begin{equation}
\label{eq36}
R_{\tilde{W}}(\mathbf{\tilde{f}}^k)=\sum_{ i,j=1,\; i<j }^{N_k} w_{i,j}^k \left(\tilde{f}^k (\mathbf{x}_i^k)-\tilde{f}^k (\mathbf{x}_j^k)  \right)^2,
\end{equation}

\noindent where $w_{i,j}^k$ is the edge weight between $i$-th and $j$-th samples in the data adjacency graph of modality $k$. \par

As proved in Appendix \ref{appC}, the right side of equation (\ref{eq36}) can be rewritten as:

\begin{equation}
\label{eq37}
\sum_{ i,j=1,\; i<j }^{N_k} w_{i,j}^k \left(\tilde{f}^k (\mathbf{x}_i^k)-\tilde{f}^k (\mathbf{x}_j^k)  \right)^2=\mathbf{\tilde{f}}^{k^T} \mathbf{L}^k \; \mathbf{\tilde{f}}^{k}.
\end{equation}

\noindent Within modality regularization for real-valued target function $\mathbf{\tilde{f}}$, $R_{\tilde{W}}(\mathbf{\tilde{f}})$, can be extended for all $m$ modalities as follows:

\begin{equation}
\label{eq38}
R_{\tilde{W}}(\mathbf{\tilde{f}})=\langle\mathbf{\tilde{f}},\mathbf{L}\mathbf{\tilde{f}} \rangle_{\mathbb{R}^N}=\mathbf{\tilde{f}}^T\mathbf{L}\mathbf{\tilde{f}}=\sum_{k=1}^m \sum_{ i,j=1,\; i<j }^{N_k} w_{i,j}^k \left(\tilde{f}^k (\mathbf{x}_i^k)-\tilde{f}^k (\mathbf{x}_j^k)  \right)^2,
\end{equation}

\noindent where $\mathbf{L}$ is the $N\times N$ diagonal block matrix, which each $N_i \times N_i$ matrix $\mathbf{L}^i$ is in its $i$-th diagonal position.\par

By extending equation (\ref{eq38}) to vector-valued target function for $c$-class classification, the within modality regularization term in vector-valued target function $\mathbf{f}\in \mathcal{Y}^N$ is defined as follows:

\begin{equation}
\label{eq39}
R_{W}(\mathbf{f})=\langle\mathbf{f},\mathbf{M}_W\mathbf{f} \rangle_{\mathcal{Y}^N}=\mathbf{f}^T\mathbf{M}_W\mathbf{f}=\sum_{k=1}^m \sum_{ i,j=1,\; i<j }^{N_k} w_{i,j}^k \parallel\mathbf{f}^k (\mathbf{x}_i^k)-\mathbf{f}^k (\mathbf{x}_j^k)  \parallel_{\mathcal{Y}^N}^2,
\end{equation}

\noindent where $\mathbf{M}_W=\mathbf{L}\otimes \mathbf{I}_c$.

\paragraph{\textbf{e) Rewriting the problem }} \mbox{}\par\nobreak
\noindent According to equations (\ref{eq31}), (\ref{eq35}), and (\ref{eq39}), problem (\ref{eq28}) can be written as:

\begin{equation}
\label{eq40}
\mathbf{f}_{opt}=\text{arg}\min_{\mathbf{f}\in\mathcal{H}_k} \frac{1}{l}\parallel\mathbf{y}- \mathbf{Jf} \parallel_{\mathcal{Y}}^2+\gamma_A \parallel\mathbf{f}\parallel_{\mathcal{H}_k}^2+\gamma_W (\mathbf{f}^T \mathbf{M}_W  \mathbf{f})+\gamma_B (\mathbf{f}^T\mathbf{M}_B  \mathbf{f}).
\end{equation}

\paragraph{\textbf{f) Reformulating for real-valued target function}} \mbox{}\par\nobreak
\noindent Considering the classical representer theorem \citep{Scholkopf2001}, the solution of the minimization problem (\ref{eq40}) that exists in $\mathcal{H}_K$ can be expressed as a finite linear combination of kernel products. According to this theorem, the minimization problem (\ref{eq40}) for a real-valued target function $\tilde{f}^i$ given a sample $\mathbf{x}_j^i$ that is the $j$-th sample in modality $i$, has a unique solution as follows:

\begin{equation}
\label{eq41}
\tilde{f}^i(\mathbf{x}_j^i)=\sum_{t=1}^{N_i} \alpha_t^i \; k^i(\mathbf{x}_t^i,\mathbf{x}_j^i).
\end{equation}

\noindent By extending equation (\ref{eq41}) for all real-valued target functions on modality $i$:

\begin{equation}
\label{eq42}
\mathbf{\tilde{f}}^i=\mathbf{K}^i\mathbf{\upalpha}^i=\begin{bmatrix} k^i(\mathbf{x}_1^i,\mathbf{x}_1^i) & \dotsm & k^i(\mathbf{x}_1^i,\mathbf{x}_{N_i}^i) \\ \vdots & \ddots & \vdots \\ k^i(\mathbf{x}_{N_i}^i,\mathbf{x}_1^i)& \dotsm & k^i(\mathbf{x}_{N_i}^i,\mathbf{x}_{N_i}^i) \end{bmatrix}\begin{bmatrix}\alpha_1^i \\ \vdots\\ \alpha_{N_i}^i\end{bmatrix},
\end{equation}

\noindent where $k^i(\mathbf{x}_p^i,\mathbf{x}_q^i)$ is the kernel function defined in modality $i$ between every two samples $\mathbf{x}_p^i$ and $\mathbf{x}_q^i$, and $\mathbf{\upalpha}^i$ is a vector of coefficients.\par

For target function $\mathbf{\tilde{f}}$, equation (\ref{eq42}) is generalized as:

\begin{equation}
\label{eq43}
\mathbf{\tilde{f}}=\mathbf{K}\mathbf{\upalpha},
\end{equation}

\noindent where $\mathbf{\upalpha}=\lbrack {\mathbf{\upalpha}^1}^T\;... \; {\mathbf{\upalpha}^m}^T\rbrack^T$ is the weight coefficient vector and $\mathbf{K}$ is a block diagonal matrix containing $\mathbf{K}^i$ in its $i$-th diagonal block.

\paragraph{\textbf{g) Reformulating for vector-valued target function}} \mbox{}\par\nobreak
\noindent We generalize equation (\ref{eq41}) for representing vector-valued target function $\mathbf{f}$ by replacing each coefficient $\alpha_t^i$ with a column vector of coefficients $\mathbf{a}_t^i=\lbrack a_{t,1}^i,...,a_{t,c}^i \rbrack^T$ as:

\begin{equation}
\label{eq44}
\mathbf{f}=\mathbf{G}\mathbf{a},
\end{equation}

\noindent where $\mathbf{G}=\mathbf{K}\otimes \mathbf{I}_c$ is the $cN\times cN$ Gram matrix and $\mathbf{a}=\lbrack {\mathbf{a}^1}^T \;...\; {\mathbf{a}^m}^T \rbrack ^T$ is the column vector with $cN$ elements.\par

By substituting (\ref{eq44}) in (\ref{eq40}), the main problem is reduced to minimizing a convex differentiable objective function over the finite-dimensional space of coefficients as follows:

\begin{equation}
\label{eq45}
  \mathbf{a}_{opt}=\text{arg}\min_{\mathbf{a}\in\mathcal{H}_k} \frac{1}{l}\ \parallel\mathbf{y}- \mathbf{JGa}\parallel_{\mathcal{Y}}^2+\gamma_A \parallel\mathbf{Ga}\parallel_{\mathcal{H}_k}^2+\gamma_W\mathbf{a}^T\mathbf{G}^T\mathbf{M}_W\mathbf{Ga}+\gamma_B\mathbf{a}^T\mathbf{G}^T\mathbf{M}_B\mathbf{Ga}.
\end{equation}

\paragraph{\textbf{h) Optimization}} \mbox{}\par\nobreak
The optimal solution of the objective function of problem (\ref{eq45}) is found by computing its derivative with respect to a and solving the equation obtained when it equals zero, as follows:

\begin{equation}
\label{eq46}
  - \frac{2}{l}\ \mathbf{JG}(\mathbf{y}-\mathbf{JGa}) +2\gamma_A \mathbf{Ga}+\gamma_W\mathbf{G}^T(\mathbf{M}_W^T+\mathbf{M}_W)\mathbf{Ga}+\gamma_B\mathbf{G}^T(\mathbf{M}_B^T+\mathbf{M}_B)\mathbf{Ga}=0,
\end{equation}

\noindent which is equivalent to:

\begin{equation}
\label{eq47}
  -\mathbf{y}+ \mathbf{JG}\mathbf{a} +2l\gamma_A \mathbf{I}\mathbf{a}+\gamma_W\frac{1}{2}(\mathbf{M}_W^T+\mathbf{M}_W)\mathbf{G}\mathbf{a}+\gamma_B\frac{1}{2}(\mathbf{M}_B^T+\mathbf{M}_B)\mathbf{G}\mathbf{a}=0.
\end{equation}

\noindent Consequently, we obtain:

\begin{equation}
\label{eq48}
  \mathbf{y}=\left( \mathbf{JG} +2l\gamma_A \mathbf{I}+\gamma_W\frac{1}{2}(\mathbf{M}_W^T+\mathbf{M}_W)\mathbf{G}+\gamma_B\frac{1}{2}(\mathbf{M}_B^T+\mathbf{M}_B)\mathbf{G}\right)\mathbf{a}.
\end{equation}

\noindent By defining $\mathbf{B}=\mathbf{JG} +2l\gamma_A \mathbf{I}+\gamma_W\frac{1}{2}(\mathbf{M}_W^T+\mathbf{M}_W)\mathbf{G}+\gamma_B\frac{1}{2}(\mathbf{M}_B^T+\mathbf{M}_B)\mathbf{G}$, the optimal coefficient vector $\mathbf{a}_{opt}$ can be obtained as follows:

\begin{equation}
\label{eq49}
  \mathbf{a}_{opt}=\mathbf{B}^{-1}\mathbf{y}.
\end{equation}

\subsubsection{Evaluation on testing samples}
\label{4.2.2EvaluationOnTestingSamples}
After coefficient vector $\mathbf{a}$ obtained based on (\ref{eq49}), it can be used for evaluation on a testing set. Evaluating on $\tilde{N}_i$ testing samples $\mathbf{V}^i=\left\{ \mathbf{v}_1^i,...,\mathbf{v}_{\tilde{N}_i}^i\right\}$ in modality $i$ is as follows:

\begin{equation}
\label{eq50}
\mathbf{f}^i=\mathbf{\tilde{G}}^i\mathbf{a}^i,
\end{equation}

\noindent where $\mathbf{f}^i=\lbrack \mathbf{f}^i (\mathbf{v}_1^i)\;...\; \mathbf{f}^i(\mathbf{v}_{\tilde{N}_i}^i)\rbrack^T$ and $\mathbf{\tilde{G}}^i=\mathbf{\tilde{K}}^i \otimes \mathbf{I}_c$ is the $c\tilde{N}_i\times c\tilde{N}_i$ Gram matrix, which $\mathbf{\tilde{K}}^i$ is $\tilde{N}_i\times N_i$ kernel matrix between testing samples and samples in modality $i$ ($k^i (\mathbf{v}_j^i,\mathbf{x}_t^i$) is the kernel function between $j$-th testing sample and $t$-th training sample in modality $i$). \par

According to our definition at the beginning of section \ref{4.2M2CPC} for representing class label in a multiclass problem, the index of the element with the maximum value in vector-valued function $\mathbf{f}^i(\mathbf{v}_j^i)$ is equivalent to the class label of $j$-th testing sample of modality $i$, $\mathbf{C}(\mathbf{v}_j^i)$, as follows:

\begin{equation}
\label{eq51}
\mathbf{C}(\mathbf{v}_j^i)=\text{arg}\max_{1\leq k \leq c} f_k^i(\mathbf{v}_j^i),
\end{equation}

\noindent where $f_k^i(\mathbf{v}_j^i)$ is the $k$-th element of vector-valued function $\mathbf{f}^i(\mathbf{v}_j^i)$.

\subsubsection{Multimodal classification algorithm}
\label{4.2.3MultimodalClassificationAlgorithm}
All the mentioned steps of the proposed multimodal classification method are summarized in Algorithm \ref{alg2}.

\begin{algorithm}[h]
  \caption{Manifold regularized multimodal classification based on pointwise correspondences (M$^2$CPC)}
  \label{alg2}
  \begin{algorithmic}[h]
    \State \textbf{Input:}
    \begin{itemize} [label=$-$]
        \item $m$: Number of modalities
        \item $\mathbf{z}=\cup_{i=1}^m \mathbf{z}^i$ : Training data, where $\mathbf{z}^i=\mathbf{z}_l^i\cup \mathbf{z}_u^i$, which $\mathbf{z}_l^i=\left\{(\mathbf{x}_j^i,\mathbf{y}_j^i)\right\}_{j=1}^{l_i}$ and $\mathbf{z}_u^i=\left\{ \mathbf{x}_j^i \right\}_{j=l_i+1}^{N_i}$ as labeled and unlabeled data in modality $i$, respectively.
        \item  $\mathbf{Y}=\lbrack\mathbf{y}^1\; ... \; \mathbf{y}^m\rbrack$: Label vectors
        \item  $\mathbf{V}^i=\left\{ \mathbf{v}_1^i,...,\mathbf{v}_{\tilde{N}_i}^i\right\}$: Testing data for each modality $i$.
        \item  $\mathbf{P}^{i,j}$: Pointwise correspondences matrices between each pairs of modalities $i$ and $j$
    \end{itemize}
    \State \textbf{Hyper-parameters:}
    \begin{itemize} [label=$-$]
        \item  $\gamma_A$, $\gamma_B$, $\gamma_W$: Regularization parameters
        \item  $k(\cdot,\cdot)$: Type of kernel function
    \end{itemize}
    \State \textbf{Procedure:}
    \begin{enumerate}
         \item Compute the geometric descriptors of all modalities and extract the Laplacian matrix of all modalities $\mathbf{L}^i$’s
         \item Compute the kernel matrix of all modalities $\mathbf{K}^i$’s
         \item Solve equation (\ref{eq49}) to obtain coefficient vector  $\mathbf{a}_{opt}=\lbrack {\mathbf{a}^1}^T\;...\; {\mathbf{a}^m}^T\rbrack^T$
    \end{enumerate}
    \State \textbf{End}
    \State \textbf{Output:}
    \begin{itemize} [label=$-$]
    \item	Evaluating all $\mathbf{\tilde{N}}_i$ testing samples $\mathbf{V}^i$ on each modality $i$ using (\ref{eq50}) and return theirs class labels using (\ref{eq51})
    \end{itemize}
  \end{algorithmic}
\end{algorithm}

\section{Experimental results}
\label{5ExperimentalResults}
In this section, we investigate the effectiveness and efficiency of our two proposed methods in cross-modal retrieval problems and multimodal multiclass classification ones on several benchmark datasets. In the following subsections, we give a brief description of the datasets used for the validation of each method, followed by an introduction of evaluation metrics. Finally, a comparison of experimental results with various state-of-the-art related methods is presented to prove the proper performance of our proposed methods.

\subsection{Datasets}
\label{5.1Datasets}
We consider two categories of datasets for cross-modal retrieval and multimodal classification problems, respectively.

\paragraph{\textbf{Cross-modal retrieval datasets}} \mbox{}\par\nobreak

\begin{itemize}
  \item \textbf{Wiki} dataset \citep{Rasiwasia2010} is generated from “Wikipedia articles” consists of $2866$ image-text pair of documents grouped into $10$ semantic categories. Image modality is described by $128$-dimensional SIFT features, and text modality is represented by $10$-dimensional topic vectors. Following the experimental protocol on state-of-the-art methods, $2173$ documents as the training set and $693$ documents as the test set are randomly selected.
  \item \textbf{Pascal VOC} dataset \citep{Sharma2012}consists of $2808/2841$ (training/testing) image-tag pairs of $20$ different categories. The image and text modalities represented by $512$-dimensional Gist features \citep{Hwang2012}, and $399$-dimensional word frequency features, respectively.
  \item \textbf{NUS-WIDE} dataset \citep{Chua2009} is a multimodal dataset containing real-world images associated with several textual tags, which are manually labeled with one or more of $81$ concept labels. Image and text modalities described by $500$-dimensional SIFT feature vectors \citep{Lowe2004}, and $1000$-dimensional tag occurrence feature vectors, respectively. The $10$ most frequent concepts containing $186,577$ samples with multiple semantic labels for each of them are selected, which among them $1000$ images with their tags were randomly chosen to serve as the query set and the remaining pairs used in the training set.
\end{itemize}

\paragraph{\textbf{Multimodal classification datasets}} \mbox{}\par\nobreak
\begin{itemize}
    \item \textbf{Caltech} dataset is a subset of the Caltech-$101$ dataset \citep{Li2006} with the same seven image classes as \citep{Cai2011} and contains two modalities. The first modality has images represented by bio-inspired features, and their annotations in the $4\times4$ pyramid histogram of visual words (PHOW) constitutes the second modality.
    \item \textbf{NUS} dataset is a subset of the NUS-WIDE dataset, which described above for cross-modal retrieval. For a fair comparison, it contains two modalities, images and their tags, as mentioned in \citep{Behmanesh2021}.
\end{itemize}

\subsection{Evaluation metrics}
\label{5.2EvaluationMetrics}
To evaluate the efficiency of the proposed method for cross-modal retrieval, two cross-modal retrieval tasks are conducted: 1) Image modality as query vs. Text modality, and 2) Text modality as query vs. Image modality. Each scenario searches relevant samples of the query in a modality (source) to another modality (target).\par

Similar to most of the existing cross-modal retrieval studies, the mean average precision (MAP) is used to evaluate the overall performance of the considered algorithms \citep{Wang2016ACS}.\par

Also, to evaluate the performance of the proposed method for multimodal classification, we use the average of the accuracy metric in $10$ runs along with their standard deviations as a standard criterion used in classification problems. The accuracy is simply measuring the ratio between the number of correct predictions and the total number of predictions.

\subsection{Performance of FMBSD method on cross-modal retrieval problems}
\label{5.3PerformanceOfFMBSDMethodOnCross-modalRetrievalProblems}
We examine the effectiveness of the proposed FMBSD method for cross-modal retrieval using two types of datasets: single-label and multi-label. Wiki and Pascal VOC are the single-label datasets, and NUS-WIDE is the multi-label one used in these experiments. In a multi-label dataset, two samples are regarded as relevant, only if they share at least one common label values.

\subsubsection{FMBSD hyper-parameter tuning}
\label{5.3.1FMBSDHyperParameterTuning}
We used Mexican hat as kernel generating function $g(\cdot)$ and a scaling function $h(\cdot)$ as suggested in \citep{HAMMOND2011}. Also, other hyperparameters, mentioned in Algorithm \ref{alg1}, are chosen through experimental verification as $R=60$, $k_i=60$, and $k=5$.\par
Since the solution quality of the problem (\ref{eq22}) depends on four hyper-parameters $\alpha$, $\beta$, $\gamma_W$, and $\gamma_B$, we tune these parameters from $\left\{10^p\right\}_{p=-3}^5$ by cross validation. The best performances of our proposed method on all datasets are obtained by setting $\alpha$, $\beta$, $\gamma_W$, and $\gamma_B$ as $10^{-1}$, $1$, $10^4$, and $10^4$ , respectively.

\subsubsection{Comparison methods}
\label{5.3.2ComparisonMethods}
To evaluate the performance of the proposed FMBSD method, we compare its experimental results with the results of several state-of-the-art methods of different types developed for realistic problems in practical scenario, containing WMCA \citep{Lampert2010}, MMPDL \citep{Huaping2018}, FlexCMH \citep{9223723}, PMH \citep{Wang2015LearningTH}, GSPH \citep{Mandal2019} and UCMH \citep{GAO2020178}, which are without pointwise correspondences (unpaired methods). The first two methods use latent subspace approaches across multimodal data, while the next methods are hashing-based ones.\par

For further investigation, we compare the performances of several methods that have a little prior knowledge containing a few pointwise or batch correspondences, which are called weakly paired correspondences, such as CCA \citep{Rasiwasia2010}, PLS \citep{Sharma2011}, FSH \citep{Liu2017}, CRE \citep{Mengqiu2019}, JFSSL \citep{Wang2016}, and DDL \citep{LIU2020199}.\par

In these experiments, the MAP score (c.f. section \ref{5.2EvaluationMetrics}) of different considered methods are compared with that of the proposed method based on the results reported by the researchers in their articles. It is clear that the comparison is only fair with methods that have no prior knowledge (neither pointwise nor batch correspondences), similar to our FMBSD method.

\subsubsection{Results on Wiki}
\label{5.3.3ResultsOnWiki}
Table \ref{table1} reports the MAP score on Wiki dataset, as the most common dataset used for evaluation of the cross-modal retrieval methods.\par

As it can be seen, the MAP score of our FMBSD method is better than all state-of-the-art unpaired methods.\par

Although FMBSD method considers a more practical scenario, which does not require any prior correspondences knowledge, it even achieves satisfactory performance compared to the JFSSL, PLS, BLM, CCA, and DDL that have weakly paired correspondences.\par

These results confirm that the superiority of FMBSD is because of providing richer knowledge about modalities by exploiting their underlying geometric structures and topological information of data, compared to the knowledge obtained by applying only semantic labels information in the mentioned previous methods.\par

Since in this dataset, the images are not closely related to their assigned categories and two modalities have a great semantic gap \citep{Zhou2609610}, the extracted image features cannot reflect the semantic properties appropriately, which leads to inconsistencies between the image features and the semantic label information. It is sometimes difficult to find semantically similar texts by image query, which results in poor MAP score for Image-query tasks compare with Text-query tasks.\par

Since the learning procedure of our FMBSD method uses only semantic similarities based on local descriptors on various modalities and is independent of label information, it avoids the inconsistencies between data samples in the Wiki dataset and, therefore, has a better MAP score than other methods.

\begin{table}[t]
\caption{MAP on Wiki dataset}
\label{table1}
\centering
\begin{tabular}{lccc}
\hline
\multirow{2}{*}{Method} & \multicolumn{3}{c}{Map}            \\ \cline{2-4}
                        & Image query & Text query & Average \\ \cline{1-4}
JFSSL \citep{Wang2016}&0.3063&0.2275&0.2669\\
PLS \citep{Sharma2011}&0.2402&0.1633&0.2032\\
CCA \citep{Rasiwasia2010}&0.2549&0.1846&0.2198\\
DDL \citep{LIU2020199}&0.2832&0.2615&0.2724\\
\textbf{WMCA} \citep{Lampert2010}&0.1336&0.1381&0.1358\\
\textbf{MMPDL} \citep{Huaping2018}&0.2268&0.2098&0.2183\\
\textbf{FlexCMH} \citep{9223723}&0.2563&0.2501&0.2532\\
\textbf{PMH} \citep{Wang2015LearningTH}&0.1235&0.1239&0.1237\\
\textbf{GSPH} \citep{Mandal2019}&0.3112&0.4370&0.3741\\
\textbf{UCMH} \citep{GAO2020178}&0.3212&0.4705&0.3958\\
\textbf{FMBSD (proposed)}&\textbf{0.4513}&\textbf{0.4731}&\textbf{0.4622}\\ \hline
\multicolumn{4}{c}{Unpaired methods are marked in bold }
\end{tabular}
\end{table}

\subsubsection{Results on Pascal VOC }
\label{5.3.4ResultsOnPascalVOC }
Table \ref{table2} shows the MAP scores achieved by the FMBSD method and on the Pascal VOC dataset and compares it with several state-of-the-art methods.\par

In many of the compared methods, which mainly focus to learn a latent subspace, the redundancy in modalities is removed by performing Principal Component Analysis (PCA) in the original feature spaces. As mentioned in section \ref{4.1FMBSDForFindingPointwiseCorrespondences}, our FMBSD method performs dimensionality reduction on various modalities simultaneously, so it ignores the redundancy in modalities.\par
It can be observed from Table \ref{table2} that most recent methods use weakly paired correspondences as prior knowledge; however, our FMBSD method without any prior knowledge outperforms them.\par

Since FMBSD method learns a mapping matrix in a reduced dimensional space which is independent of the samples, it performs better in higher dimensional datasets. Also, our method is more effective because of exploiting between-modality and within-modality similarity relationships, which further improves the performance.

\begin{table}[t]
\caption{MAP on Pascal VOC dataset}
\label{table2}
\centering
\begin{tabular}{lccc}
\hline
\multirow{2}{*}{Method} & \multicolumn{3}{c}{Map}            \\ \cline{2-4}
                        & Image query & Text query & Average \\ \cline{1-4}
PCA+CCA-3V \citep{Rasiwasia2010}&0.3146&0.2562&0.2854\\
PCA+SCM \citep{Pereira2014OnTR}&0.3216&0.2475&0.2846\\
PCA+PLS \citep{Sharma2011}&0.2757&0.1997&0.2377\\
PCA+CCA \citep{Rasiwasia2010}&0.2655&0.2215&0.2435\\
JFSSL \citep{Wang2016}&0.3607&0.2801&0.3204\\
DDL \citep{LIU2020199}&0.3261&0.2948&0.3105\\
\textbf{FMBSD (proposed)}&\textbf{0.3825}&\textbf{0.3573}&\textbf{0.3699}\\ \hline
\multicolumn{4}{c}{Unpaired methods are marked in bold }
\end{tabular}
\end{table}

\subsubsection{Results on NUS-WIDE}
\label{5.3.5ResultsOnNUS-WIDE}

Table \ref{table3} shows the MAP scores achieved by the FMBSD method on the NUS-WIDE dataset. Since in this multi-label dataset, at least one shared common label is needed for two samples to regard them as relevant, the average of the score in most of the methods are higher than those of the previous datasets. It can be observed that FMBSD method outperforms all of its counterparts.\par

According to Table \ref{table3}, the MAP of FMBSD method is not only better than all of the state-of-the-art unpaired methods, but also is better than all recent weakly paired methods. Although it is thought that CCA, FSH, and CRE, as the methods that use prior knowledge, should provide better results, our proposed FMBSD method without any prior knowledge achieved the best performances, which confirm the prominent role of the geometric structure of data in comparison with applying a little prior knowledge.\par

Also, compared with the other unpaired methods such as FlexCMH, which uses the centroids of data clusters in each modality for extracting the local structures, our FMBSD method extracts efficient descriptors, which effectively encode the geometric structure of data and use them properly for finding a functional mapping between modalities.\par

Since the experimental settings for this dataset are quite close to the real-world scenarios, the results of Table \ref{table3} demonstrate the superior performance of our method in handling the real-world large-scale cross-modal retrieval problems.\par

It can be observed from Tables \ref{table2} and \ref{table3} that in high dimensional multimodal datasets, our proposed method outperforms all of its counterparts. It can be because that our method learns a mapping matrix in a reduced dimensional space which is independent of samples.

\begin{table}[t]
\caption{MAP on NUS-WIDE dataset}
\label{table3}
\centering
\begin{tabular}{lccc}
\hline
\multirow{2}{*}{Method} & \multicolumn{3}{c}{Map}            \\ \cline{2-4}
                        & Image query & Text query & Average \\ \cline{1-4}
CCA \citep{Rasiwasia2010}&0.3560&0.3545&0.3552\\
FSH \citep{Liu2017}&0.5171&0.4965&0.5068\\
CRE \citep{Mengqiu2019}&0.5338&0.5161&0.5249\\
\textbf{WMCA} \citep{Lampert2010}&0.3604&0.3481&0.3542\\
\textbf{MMPDL} \citep{Huaping2018}&0.4015&0.3713&0.3864\\
\textbf{FlexCMH} \citep{9223723}&0.4273&0.4212&0.4242\\
\textbf{PMH} \citep{Wang2015LearningTH}&0.3433&0.3437&0.3435\\
\textbf{GSPH} \citep{Mandal2019}&0.3895&0.4438&0.4166\\
\textbf{UCMH} \citep{GAO2020178}&0.5210&0.5266&0.5238\\
\textbf{FMBSD (proposed)}&\textbf{0.5490}&\textbf{0.5774}&\textbf{0.5632}\\ \hline
\multicolumn{4}{c}{Unpaired methods are marked in bold }
\end{tabular}
\end{table}

\subsection{Performance of M$^2$CPC method on multimodal classification problems}
\label{5.4PerformanceOfM2CPCMethod}
To verify the effectiveness of our M$^2$CPC in multimodal multiclass problems, firstly, we compare its accuracy in some of datasets with two unpaired and paired scenarios. Since most datasets are based on paired scenarios, to have unpaired datasets, we randomly shuffle the order of data samples in each modality, which makes the pointwise correspondences between the original modalities unknown.\par

Table \ref{table4} reports the means and standard deviations of accuracies in $20$ runs of M$^2$CPC method (two scenarios) and several state-of-the-art methods on two datasets. In this table, the results of the first four methods are obtained based on $20$ percent of paired samples between modalities. Although our approach has no correspondence between data samples in the unpaired scenario, it can be seen that it provides good performances.\par

We can also see that the performance of our M$^2$CPC method is significantly improved in the paired scenario, which demonstrates the better performance of the proposed manifold regularization framework for multimodal classification compared to existing methods.

\begin{table}[t]
\caption{Classification accuracy (Mean Accuracy $\pm$ Standard Deviations) }
\label{table4}
\centering
\begin{tabular}{lccc}
\hline
\multirow{2}{*}{Method} & \multicolumn{2}{c}{Datasets}            \\ \cline{2-3}
                        & Caltech & NUS \\ \cline{1-3}
CD (pos) \citep{Eynard2015}&76.8$\pm$0.7&81.4$\pm$0.5\\
CD (pos+neg) \citep{Eynard2015}&73.7$\pm$0.3&78.9$\pm$0.1\\
SCSMM \citep{POURNEMAT2021}&-&83.9 $\pm$2.42\\
m-LSEJD \citep{Behmanesh2021}&82.2$\pm$2.4&87.1$\pm$1.5\\
m$^2$-LSEJD \citep{Behmanesh2021}&89.1$\pm$0.4&89.6$\pm$0.4\\
M$^2$CPC-paired (proposed)& \textbf{89.1$\pm$0.3} & \textbf{90.2$\pm$0.2}\\
M$^2$CPC-unpaired(proposed)&84.8$\pm$0.7&86.4$\pm$0.3\\   \hline
\end{tabular}
\end{table}

\subsubsection{M$^2$CPC hyper-parameter tuning}
\label{5.4.1M2CPCHyper-parameterTuning}

The efficiency of the proposed method is moderately related to several regularization parameters involved in the proposed method, containing the RKHS regularization parameter $\gamma_A$, and manifold regularization parameters $\gamma_w$ and $\gamma_B$. \par

\begin{figure}[t!]
\centering
  \begin{tabular}[c]{lc}
    \begin{subfigure}[c]{0.5\textwidth}
      \includegraphics[width=3in]{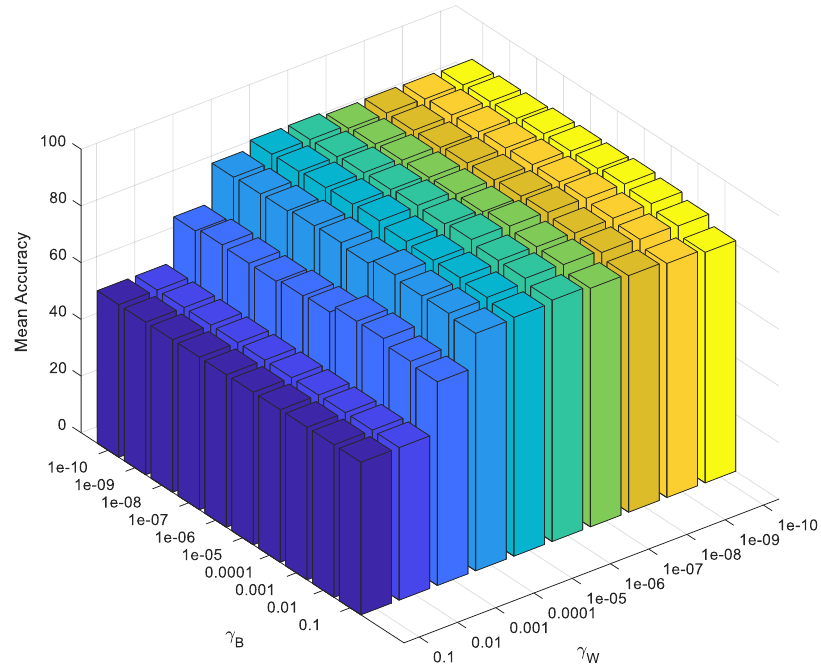}
      \caption{}
      \label{fig3a}
    \end{subfigure}&
    \begin{subfigure}[c]{0.5\textwidth}
      \includegraphics[width=3in]{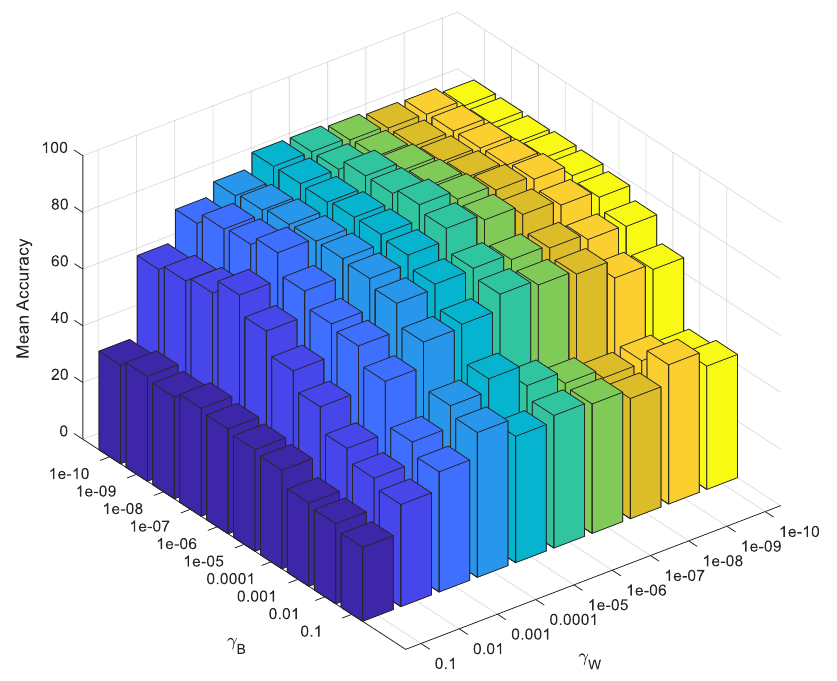}
      \caption{}
      \label{fig3b}
    \end{subfigure}\\
    \begin{subfigure}[c]{0.5\textwidth}
      \includegraphics[width=3in]{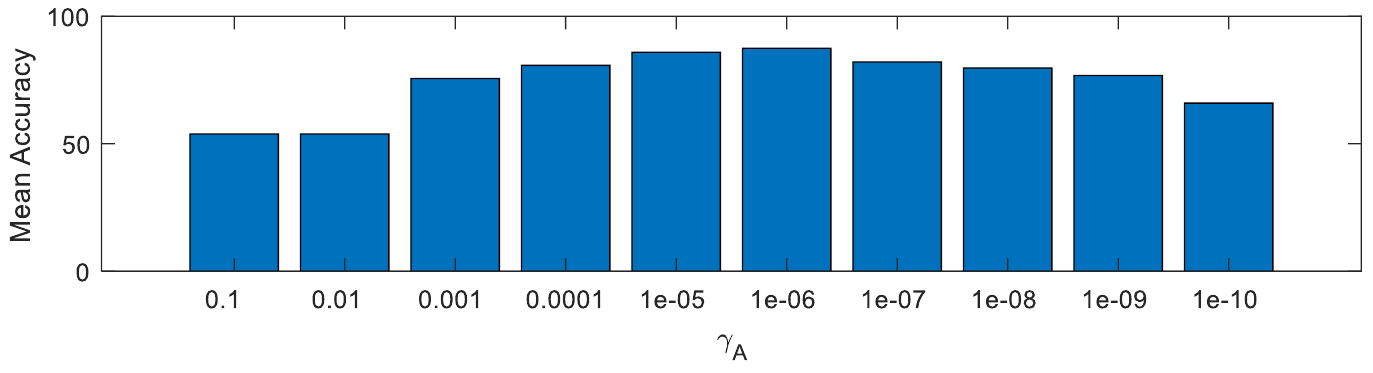}
      \caption{}
      \label{fig3c}
    \end{subfigure}&
    \begin{subfigure}[c]{0.5\textwidth}
      \includegraphics[width=3in]{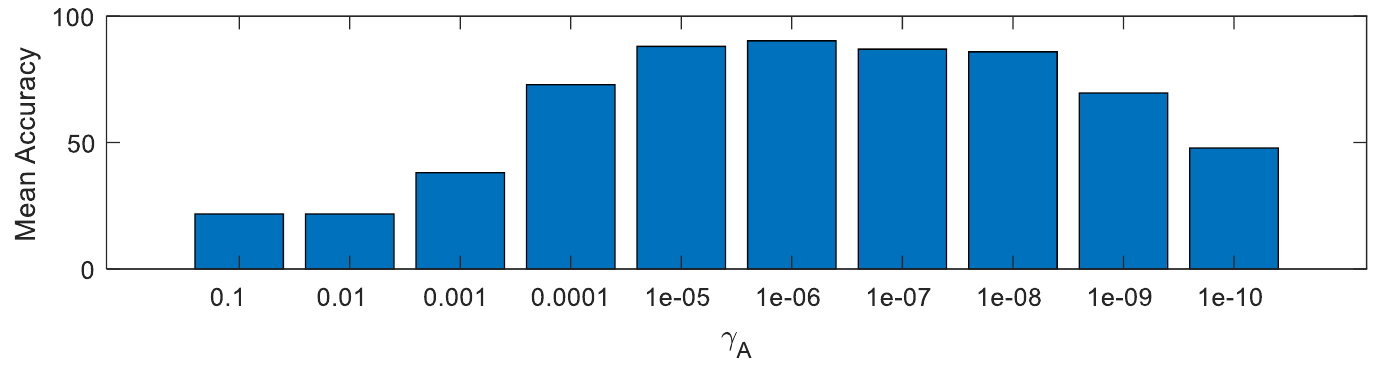}
      \caption{}
      \label{fig3d}
    \end{subfigure}\\
  \end{tabular}
  \caption{Performance evaluation by fixing $\gamma_A$ for Caltech (a), and NUS (b), and by fixing $\gamma_W$ and $\gamma_B$ for Caltech (c), and NUS (d) datasets}
\label{fig3}
\end{figure}

We tune these three parameters in the range of $\left\{10^{-p} \right\}_{p=1}^{10}$. The experimental results of the proposed M$^2$CPC method based on the paired scenario in the NUS and Caltech datasets are shown in Fig. \ref{fig3}. First, we fix $\gamma_A$ and evaluate the performance of the method by changing the manifold regularization parameters $\gamma_w$ and $\gamma_B$. The obtained results for two datasets Caltech and NUS are shown in Fig. \ref{fig3a} and Fig. \ref{fig3b}, respectively. Similarly, the results obtained by considering fixed values for $\gamma_w$ and $\gamma_B$ and changing $\gamma_A$ for two datasets Caltech and NUS are shown in Fig. \ref{fig3c} and Fig. \ref{fig3d}, respectively. As can be seen from this figure, the performance of our method more or less varies by changing the different parameters. Thus, for the NUS dataset, the proposed method gives the best performance by setting the parameters $\gamma_A$, $\gamma_w$, and $\gamma_B$ as $10^{-6}$, $10^{-6}$, and $10^{-6}$, respectively. Also, for the Caltech dataset, the proposed method obtains the best performance by setting the parameters $\gamma_A$, $\gamma_w$, and $\gamma_B$ as $10^{-6}$, $10^{-6}$, and $10^{-4}$, respectively, which are close to the values obtained for NUS. The other hyper-parameter is the type of kernel function, which in all experiments, we use Gaussian kernel.

\section{Conclusions}
\label{6.Conclusions}
In this paper, we have focused on analyzing the multimodal problems in a more practical scenario with different number of heterogeneous data samples in each modality and without any prior knowledge between modalities, i.e., pointwise correspondences or batch correspondences. Two methods have been proposed under this scenario.\par

The first method (FMBSD) is proposed as an independent model of correspondence between modalities that uses spectral graph wavelet transform for representing the localities of each manifold. A new regularization-based framework is established to find a functional map between modalities that are expected to preserve these local descriptors. This mapping is well used to find pointwise correspondences in multimodal problems and also applied in cross-modal retrieval problems.\par

Extensive experiments on three cross-modal retrieval problems demonstrate the superiority of FMBSD method over state-of-the-art methods for more practical scenario without any prior knowledge. Also, compared to methods that use pointwise correspondences as prior knowledge, the results show the superiority of FMBSD method while it does not require any prior correspondences knowledge.\par

Under the supposed scenario, the second proposed method (M$^2$CPC) introduces a new multimodal multiclass classification model. This method uses a manifold regularization framework to combine the vector-valued reproducing kernel Hilbert spaces (RKHS) for multiclass data and a pointwise correspondence framework for multimodal heterogeneous data, where the correspondences between modalities are determined based on the first model.\par

Experimental results on two benchmark multimodal multiclass datasets demonstrate that the M$^2$CPC proposed method outperforms state-of-the-art methods.




\newpage

\appendix
\section{}
\label{appA}


In this appendix, we formulate equation (\ref{eq15}) to the matrix form.

\noindent Equation (\ref{eq15}) can be written as follows:

\begin{align*}
\Omega_B(\mathbf{C}^{i,j})=&\sum_{r=1}^{N_i}\sum_{t=1}^{N_j}\pi_{r,t}^{i,j} \parallel \mathbf{\upgamma}_r^{i,j}-\mathbf{\updelta}_t^j\parallel^2\\
&=\sum_{r=1}^{N_i}\sum_{t=1}^{N_j} \parallel \mathbf{\upgamma}_r^{i,j}\parallel^2 \pi_{r,t}^{i,j} + \sum_{r=1}^{N_i}\sum_{t=1}^{N_j} \parallel \mathbf{\updelta}_t^j\parallel^2 \pi_{r,t}^{i,j}-2\sum_{r=1}^{N_i}\sum_{t=1}^{N_j} \pi_{r,t}^{i,j}{\mathbf{\upgamma}_r^{i,j}}^T\mathbf{\updelta}_t^j.
\end{align*}

\noindent By considering $d_r^{i,j}=\sum_{t=1}^{N_j} \pi_{r,t}^{i,j}$ :

\begin{align*}
\sum_{r=1}^{N_i}\sum_{t=1}^{N_j} \parallel \mathbf{\upgamma}_r^{i,j}\parallel^2 \pi_{r,t}^{i,j} =&\sum_{r=1}^{N_i} {\mathbf{\upgamma}_r^{i,j}}^T d_{r}^{i,j}\mathbf{\upgamma}_r^{i,j} \\
&=\begin{bmatrix}{\mathbf{\upgamma}_1^{i,j}}^T d_{1}^{i,j} & \dots & {\mathbf{\upgamma}_{N_i}^{i,j}}^T d_{N_i}^{i,j} \end{bmatrix}\begin{bmatrix}\upgamma_1^{i,j} \\ \vdots  \\ \upgamma_{N_i}^{i,j}  \end{bmatrix} \\
&=\begin{bmatrix}{\mathbf{\upgamma}_1^{i,j}}^T & \dots & {\mathbf{\upgamma}_{N_i}^{i,j}}^T\end{bmatrix} \begin{bmatrix}d_1^{i,j} & \dotsm & 0\\ \vdots & \ddots & \vdots \\ 0 & \dotsm & d_{N_i}^{i,j} \end{bmatrix} \begin{bmatrix}\upgamma_1^{i,j} \\ \vdots  \\ \upgamma_{N_i}^{i,j}  \end{bmatrix}  \\
&=tr\left( {\mathbf{\Gamma}^{i,j}}^T \mathbf{D}^{i}\mathbf{\Gamma}^{i,j}\right),
\end{align*}

\noindent which $\mathbf{D}^i=diag(d_1^{i,j},...,d_{N_i}^{i,j})$ is diagonal matrix of size $N_i$.\par

\noindent Similarly, $\sum_{r=1}^{N_i}\sum_{t=1}^{N_j} \parallel \mathbf{\updelta}_r^{i,j}\parallel^2 \pi_{r,t}^{i,j}
=tr\left( {\mathbf{\Delta}^{j}}^T \mathbf{\tilde{D}}^{j}\mathbf{\Delta}^{j}\right)$, where $\mathbf{\tilde{D}}^j=diag(\tilde{d}_1^{i,j},...,\tilde{d}_{N_i}^{i,j})$ and $\tilde{d}_r^{i,j}=\sum_{r=1}^{N_i} \pi_{r,t}^{i,j}$.

\noindent Also:

\begin{align*}
\sum_{r=1}^{N_i}\sum_{t=1}^{N_j} \pi_{r,t}^{i,j}{\mathbf{\upgamma}_r^{i,j}}^T\mathbf{\updelta}_t^j=& \sum_{t=1}^{N_j} \sum_{r=1}^{N_i}{\mathbf{\upgamma}_r^{i,j}}^T\pi_{r,t}^{i,j}\mathbf{\updelta}_t^j\\
& =\begin{bmatrix}\sum_{r=1}^{N_i}{\mathbf{\upgamma}_r^{i,j}}^T\pi_{r,1}^{i,j} & \dotsm & \sum_{r=1}^{N_i}{\mathbf{\upgamma}_r^{i,j}}^T\pi_{r,{N_j}}^{i,j}\end {bmatrix}\begin{bmatrix} \mathbf{\updelta}_1^j \\ \vdots \\ \mathbf{\updelta}_{N_j}^j \end{bmatrix}\\
&=\begin{bmatrix}{\mathbf{\upgamma}_1^{i,j}}^T & \dotsm & {\mathbf{\upgamma}_{N_i}^{i,j}}^T\end {bmatrix}\begin{bmatrix}\pi_{1,1}^{i,j} & \dotsm & \pi_{1,N_j}^{i,j}\\ \vdots & \ddots & \vdots\\ \pi_{N_i,1}^{i,j} & \dotsm & \pi_{N_i,N_j}^{i,j}\end{bmatrix}\begin{bmatrix} \mathbf{\updelta}_1^j \\ \vdots \\ \mathbf{\updelta}_{N_j}^j \end{bmatrix}\\
&=tr\left( {{\mathbf{\Gamma}}^{i,j}}^T \mathbf{\Pi}^{i,j}\mathbf{\Delta}^{j}\right).
\end{align*}

\noindent Thus, equation (\ref{eq16}) is equivalent to equation (\ref{eq15}) as follows:

\begin{align*}
  \Omega_B(\mathbf{C}^{i,j})=tr\left({\mathbf{\Gamma}^{i,j}}^T \mathbf{D}^i \mathbf{\Gamma}^{i,j}-2{\mathbf{\Gamma}^{i,j}}^T\mathbf{\Pi}^{i,j}\mathbf{\Delta}^{j}+{\mathbf{\Delta}^{j}}^T\tilde{\mathbf{D}}^{j}\mathbf{\Delta}^{j}\right).
\end{align*}

\section{}
\label{appB}


In this appendix, we compute the derivative of each term of problem (\ref{eq22}). \\

\noindent \textbf{1) Derivative of objective function:}\par
\noindent Derivative of the first term of equation (\ref{eq13}) with respect to $\mathbf{C}^{i,j}$ is as follows:

\begin{align*}
\triangledown_{\mathbf{C}^{i,j}}&\left(\parallel {\mathbf{C}^{i,j}}^T\mathbf{A}^i-\mathbf{A}^j \parallel_F^2 \right)=\triangledown_{\mathbf{C}^{i,j}}\; tr \left[\left( {\mathbf{C}^{i,j}}^T\mathbf{A}^i-\mathbf{A}^j \right)\left( {\mathbf{C}^{i,j}}^T\mathbf{A}^i-\mathbf{A}^j \right)^T \right]\\
&=\triangledown_{\mathbf{C}^{i,j}}\; tr \left[ {\mathbf{C}^{i,j}}^T\mathbf{A}^i{\mathbf{A}^i}^T \mathbf{C}^{i,j}-{\mathbf{C}^{i,j}}^T\mathbf{A}^i{\mathbf{A}^j}^T-\mathbf{A}^j{\mathbf{A}^i}^T \mathbf{C}^{i,j}+\mathbf{A}^j{\mathbf{A}^j}^T\right]\\
&=2\mathbf{A}^i{\mathbf{A}^i}^T\mathbf{C}^{i,j}-2\mathbf{A}^i{\mathbf{A}^j}^T=2\mathbf{A}^i\left({\mathbf{C}^{i,j}}^T\mathbf{A}^i-\mathbf{A}^j \right)^T.
\end{align*}

\noindent Derivative of the second term of equation (\ref{eq13}) with respect to $\mathbf{C}^{i,j}$ is as follows:

\begin{align*}
\triangledown_{\mathbf{C}^{i,j}}&\left( \sum_{k=1}^q   \parallel \mathbf{\Phi}_k^i\mathbf{C}^{i,j}-\mathbf{C}^{i,j}\mathbf{\Phi}_k^j \parallel_F^2 \right)\\
&=2 \sum_{k=1}^q  {\mathbf{\Phi}_k^i}^T \left(\mathbf{\Phi}_k^i\mathbf{C}^{i,j}-\mathbf{C}^{i,j}\mathbf{\Phi}_k^j \right)-2 \sum_{k=1}^q   \left(\mathbf{\Phi}_k^i\mathbf{C}^{i,j}-\mathbf{C}^{i,j}\mathbf{\Phi}_k^j \right){\mathbf{\Phi}_k^j}^T.
\end{align*}

\noindent \textbf{2) Derivative of manifold regularization term:}\par

\noindent Derivative of the first term of equation (\ref{eq20}) with respect to $\mathbf{C}^{i,j}$ is as follows:

\begin{align*}
\triangledown_{\mathbf{C}^{i,j}}& \left(  tr\left({\mathbf{C}^{i,j}}^T {\mathbf{\Delta}^{i}}^T \mathbf{D}^{i}\mathbf{\Delta}^{i}\mathbf{C}^{i,j}-2{\mathbf{C}^{i,j}}^T {\mathbf{\Delta}^{i}}^T \mathbf{\Pi}^{i,j}\mathbf{\Delta}^{j}+{\mathbf{\Delta}^{j}}^T \tilde{\mathbf{D}}^{j}\mathbf{\Delta}^{j}\right) \right)\\
&={\mathbf{\Delta}^{i}}^T \mathbf{D}^{i}\mathbf{\Delta}^{i}\mathbf{C}^{i,j}+{\mathbf{\Delta}^{i}}^T {\mathbf{D}^{i}}^T\mathbf{\Delta}^{i}\mathbf{C}^{i,j}-2{\mathbf{\Delta}^{i}}^T \mathbf{\Pi}^{i,j}\mathbf{\Delta}^{j}\\
&=2{\mathbf{\Delta}^{i}}^T \mathbf{D}^{i}\mathbf{\Delta}^{i}\mathbf{C}^{i,j}-2{\mathbf{\Delta}^{i}}^T \mathbf{\Pi}^{i,j}\mathbf{\Delta}^{j}.
\end{align*}

\noindent Derivative of the second term of equation (\ref{eq20}) with respect to $\mathbf{C}^{i,j}$ is as follows:

\begin{align*}
\triangledown_{\mathbf{C}^{i,j}}\left(   tr\left({\mathbf{C}^{i,j}}^T{\mathbf{\Delta}^{i}}^T\mathbf{L}^{i} \mathbf{\Delta}^{i}\mathbf{C}^{i,j}\right)\right)=2{\mathbf{\Delta}^{i}}^T\mathbf{L}^{i}\mathbf{\Delta}^{i}\mathbf{C}^{i,j}.
\end{align*}

\section{}
\label{appC}
In this appendix, we formulate equation (\ref{eq36}) to the matrix form. \par

\noindent The right side of equation (\ref{eq36}) can be written as follows:

\begin{align*}
\frac{1}{2}\sum_{i,j=1,\;i<j}^{N} w_{i,j}&\left( f(\mathbf{x}_i)-f(\mathbf{x}_j)\right)^2 =\frac{1}{2}\left( \sum_{i=1}^N d_i f(\mathbf{x}_i)^2-2\sum_{i,j=1}^N f(\mathbf{x}_i)f(\mathbf{x}_j)w_{i,j}+\sum_{j=1}^N d_j f(\mathbf{x}_j)^2\right)\\
&=\sum_{i=1}^N d_i f(\mathbf{x}_i)^2-\sum_{i,j=1}^N f(\mathbf{x}_i)f(\mathbf{x}_j)w_{i,j}=\mathbf{f}^T\mathbf{D} \mathbf{f}-\mathbf{f}^T\mathbf{W} \mathbf{f}=\mathbf{f}^T\mathbf{L} \mathbf{f},
\end{align*}

\noindent where $d_i=\sum_{j=1}^N w_{i,j}$ and $\mathbf{D}=daig(d_1,...d_N)$.

\vskip 0.2in
\bibliography{paper2}

\end{document}